\documentclass[twoside]{article}
\usepackage[accepted]{aistats2017}

\usepackage{hyperref}
\usepackage[caption=false]{subfig}
\usepackage{multirow}
\usepackage{amsmath}
\usepackage{natbib}
\usepackage{algorithm}
\usepackage{mathtools}
\usepackage{Definitions}
\usepackage{color}
\usepackage{hhline}
\usepackage{pifont}

\newcommand{\cmark}{\ding{51}}
\newcommand{\xmark}{\ding{55}}

\newcommand{\xhdr}[1]{\vspace{0mm}\noindent{{\bf #1.}}}

\begin{document}

\twocolumn[

\aistatstitle{Fairness Constraints: Mechanisms for Fair Classification}
\aistatsauthor{ Muhammad Bilal Zafar, Isabel Valera, Manuel Gomez Rodriguez, Krishna P. Gummadi }
\aistatsaddress{ Max Planck Institute for Software Systems (MPI-SWS), Germany }

]

\vspace{-3mm}
\begin{abstract}
\vspace{-3mm}
Algorithmic decision making systems are ubiquitous across a wide variety of online as well as offline services.
These systems rely on complex learning methods and vast amounts of data to optimize the service functionality, satisfaction
of the end user and profitability.
However, there is a growing concern that these automated decisions can lead,
even in the absence of intent,
to a lack of fairness, \ie, their outcomes can disproportionately hurt (or, benefit)
particular groups of people
sharing one or more sensitive attributes (\eg, race, sex).
In this paper, we introduce a flexible mechanism to design fair classifiers
  by leveraging a novel
 intuitive measure of decision boundary (un)fairness.
We instantiate this mechanism with two well-known classifiers, logistic regression and support vector machines, and show on real-world data
that our mechanism allows for a fine-grained control on the degree of fairness, often at a small cost in terms of accuracy.
\\
A Python implementation of our mechanism is available at \textbf{\url{fate-computing.mpi-sws.org}}

\end{abstract}

\vspace{-5mm}
\section {INTRODUCTION} \label{sec:intro}
\vspace{-3mm}

Algorithmic decision making processes are increasingly becoming automated and data-driven in both online (\eg, spam filtering, product personalization)
as well as offline (\eg, pretrial risk assessment, mortgage approvals) settings.
However, as automated data analysis replaces human supervision in decision making, and the scale of the analyzed
data becomes ``big'', there are growing concerns from civil organizations~\citep{bhandari_aclu}, governments~\citep{bigdatawhitehouse2014,bigdatawhitehouse2016},
and researchers~\citep{sweeney_queue} about potential loss of transparency, accoun\-ta\-bi\-li\-ty and fairness.

Anti-discrimination laws in many countries prohibit unfair treatment of people based on certain attributes, also called sensitive attributes, such as gender or race~\citep{civil_rights_act}.
These laws typically evaluate the fairness of a decision making process by means of two distinct notions~\citep{barocas_2016}:
\textbf{\textit{disparate treatment}}
and
\textbf{\textit{disparate impact}}.
A decision making
process suffers from disparate treatment if its decisions are (partly) based on the subject's sensitive attribute information,
and it has disparate impact if its {outcomes} disproportionately hurt (or, benefit) people with certain sensitive attribute va\-lues (\eg, females, blacks).

While it is desirable to design decision making systems free of disparate treatment as well as disparate impact,
controlling for both
forms of unfairness \textit{simultaneously} is challenging. One could avoid disparate treatment by
ensuring that the decision making process does not have access to sensitive attribute information (and hence cannot make use of it).
However, ignoring the sensitive attribute information may still lead to disparate impact in outcomes:
since automated decision-making systems are often trained on historical data, if a group with a certain sensitive attribute value was unfairly treated in the past,
\footnote{Like earlier studies on fairness aware-learning, we assume that while historical class labels may be biased
against group(s) with certain sensitive attribute value(s), they still contain \textit{some} degree of information on the \textit{true} (unbiased) labels. Assuming the class labels to be completely biased and having no information on the true labels would render a learning task on such a dataset pointless.}
this unfairness may persist in future predictions through \textit{indirect discrimination}~\citep{pedreschi_discrimination}, leading to disparate impact.
Similarly, avoiding disparate impact in outcomes by using sensitive attribute information while making decisions would constitute
disparate treatment, and may also lead to \textit{reverse discrimination}~\citep{ricci_scotus}.

In this work, our goal is to design classifiers---specifically, convex margin-based classifiers like logistic regression and support vector machines (SVMs)---that avoid \textit{both} disparate treatment and disparate impact,
and can additionally accommodate the ``business necessity'' clause of disparate impact doctrine.
According to the business necessity clause, an employer can justify a certain degree of disparate impact in order to meet certain performance-related constraints
~\citep{barocas_2016}. However, the employer needs to ensure that the current decision making incurs the \textit{least possible} disparate impact under the given constraints.
To the best of our knowledge, this clause has not been addre\-ssed by
any prior study on fairness-aware lear\-ning.

While there is no specific numerical formula laid out by anti-discrimination laws to quantify disparate impact, here we leverage a specific instantiation supported by the U.S. Equal Employment Opportunity Commission: the \textit{``80\%-rule''} (or more generally, the $p$\%-rule)~\citep{2005adverse}. The $p$\%-rule states that the ratio between the percentage of
subjects ha\-ving a certain sensitive attribute value assigned the positive decision outcome and the percentage of subjects not having that value also assigned the positive outcome should be no less than p:100.
Since it is very challenging to directly incorporate this rule in the formulation of convex margin-based classifiers,
we introduce a novel intuitive measure of decision boundary (un)fairness as a tractable proxy to the rule:
the covariance between the sensitive attri\-butes and the (signed) distance
 between the subjects'{} feature vectors and the decision boundary of the classifier.

Our measure  of fairness allows us to derive two complementary formulations for training fair classifiers: one that maximizes accuracy subject to fairness constraints, and enables compliance with disparate impact doctrine in its basic form (\ie, the $p$\%-rule); and another that maximizes fairness subject to accuracy constraints, and ensures fulfilling the business necessity clause of disparate impact.
Remarkably, both formulations also avoid disparate treatment, since they do not make use of sensitive attribute information
while making decisions.
Our measure additionally satisfies several desirable properties:
(i) for a wide variety of convex margin-based (linear and non-linear) classifiers, it is convex and can be readily incorporated in their formulation
without increasing their complexity;
(ii) it allows for clear mechanism to trade-off fairness and accuracy; and,
(iii) it can be used to ensure fairness with respect to several sensitive attributes.
Experiments with two well-known classifiers, logistic regression and support vector machines, using both
synthetic and real-world data show that our fairness measure allows for a fine-grained control of the level of fairness, often at
a small cost in terms of accuracy, and provides more flexibility than the state-of-the-art (see Table~\ref{table:method_comparison}).

\begin{table*}
\centering
\resizebox{\textwidth}{!}{
\begin{tabular}{l||c|c|c|c|c|c}
\textbf{Method} & \textbf{\shortstack{Disp.\\Treat.}} & \textbf{\shortstack{Disp.\\Imp.}} &  \textbf{\shortstack{Business.\\Necessity}} &  \textbf{\shortstack{Polyvalent\\sens. attrs.}} &  \textbf{\shortstack{Multiple\\sens. attrs.}} &  \textbf{\shortstack{Range of\\classifiers}}  \\
\hline
\hline
Our method                      & \cmark   & \cmark  &  \cmark & \cmark & \cmark & Any convex margin-based \\ \hline
\cite{kamiran_sampling}         & \cmark   & \cmark  &  \xmark & \xmark & \xmark & Any score-based \\ \hline
\cite{cadlers_naivebayes}       & \cmark   & \cmark  &  \xmark & \xmark & \xmark & Naive Bayes \\ \hline
\cite{salvatore_knn}       		& \cmark   & \xmark  &  \xmark & \xmark & \xmark & Any score-based \\ \hline
\cite{kamishima_regularizer}    & \xmark   & \cmark  &  \xmark & \xmark & \xmark & Logistic Regression \\ \hline
\cite{icml2013_zemel13}         & \cmark   & \cmark  &  \xmark & \xmark & \xmark & Log loss \\ \hline
\cite{feldman_kdd15}            & \cmark   & \cmark  &  \xmark & \cmark & \cmark & Any {\tiny (only numerical non-sens. attrs.)} \\ \hline
\cite{goh_nips2016}             & \cmark   & \cmark  &  \xmark & \cmark & \cmark & Ramp loss \\ \hline
\end{tabular}
}
\vspace{-3mm}
\caption{Capabilities of different methods in eliminating disparate impact and/or disparate treatment. None of the prior methods addresses disparate impact's business necessity clause. Many of the methods do not generalize to multiple (\eg, gender and race) or polyvalent sensitive attributes (\eg, race, that has more than two values).}
\label{table:method_comparison}
\vspace{-3mm}
\end{table*}

\xhdr{Related Work}
A number of prior studies have focused on controlling disparate impact and/or disparate treatment-based discrimination in the context of binary classification~\citep{salvatore_survey}.
These studies have typically adopted one of the two following strategies:

The first strategy consists of pre-processing the training data~\citep{Dwork2012,feldman_kdd15,kamiran_classifying,kamiran_sampling}.
This typically involves (i) changing the value of the sensitive attributes or class labels of individual items in the training data, or (ii) mapping the training data to a transformed
space where the dependencies between sensitive attri\-butes and class labels disappear.
However, these approaches treat the learning algorithm as a \emph{black box} and, as a consequence, the pre-processing can lead to unpredictable losses in accuracy.

The second strategy consists of modifying existing classifiers to limit discrimination~\citep{cadlers_naivebayes, kamishima_regularizer, goh_nips2016}.
Among them, the work by Kamishima et al. ~\citep{kamishima_regularizer} is the most closely related to ours: it introduces a re\-gu\-lari\-zation term to penalize discrimination in the formulation of
the logistic regression classifier.

Recently,~\citet{icml2013_zemel13}, building on~\cite{Dwork2012}, combined both strategies by jointly learning a fair representation of the data and the classifier pa\-ra\-me\-ters.
This approach has two main limitations:
i) it leads to a non-convex optimization problem and does not guarantee optimality, and ii) the accuracy of the classifier depends
on the dimension of the fair representation, which needs to be chosen rather arbitrarily.

Many of the prior studies suffer from one or more of the following limitations:
(i) they are restricted to a narrow range of classifiers,
(ii) they only accommodate a single, binary sensitive attribute,
and (iii) they cannot eliminate disparate treatment and disparate impact \textit{simultaneously}.
Table~\ref{table:method_comparison} compares the capabilities of different methods while achieving fairness.

Finally, as discussed earlier, disparate impact is particularly well suited as a fairness criterion when his\-to\-ri\-cal decisions used during the training phase are biased against
certain social groups. In such contexts, pro\-por\-tio\-na\-li\-ty in outcomes (\eg, $p$\%-rule) may help mitigate these historical biases.
However, in cases where the (unbiased) ground-truth is available for the training phase, \ie, one can tell whether a historical decision was \emph{right} or \emph{wrong},
disproportionality in outcomes can be explained by the means of the ground-truth.
In those cases, disparate impact may be a rather misleading notion of fairness, and other recently proposed criteria like ``disparate mistreatment'' by~\citet{zafar_dmt},
may be better suited notions of fairness. For more discussion into this alternative notion, we point the reader to our companion paper~\citep{zafar_dmt}.

\vspace{-4mm}
\section {{FAIRNESS IN CLASSIFICATION}}
\vspace{-4mm}

For simplicity, we consider binary classification tasks in this work. However, our ideas can be easily extended to m-ary classification.

In a binary classification task,
one needs to find a ma\-pping function $f(\mathbf{x})$ between user feature vectors $\mathbf{x} \in \RR^d$ and class labels $y \in \{-1, 1\}$. This task is achieved by utilizing a trai\-ning set, $\{ (\mathbf{x}_i, y_i) \}_{i=1}^{N}$, to construct a mapping that works \emph{well}
on an \emph{unseen} test
set.
For margin-based classifiers, finding this mapping usually reduces to building a decision boundary in feature space that separates users in the training set accor\-ding to their class
labels.
One typically looks for a decision boundary, defined by a set of parameters $\thetab^{*}$, that achieves the greatest classification accuracy in a test set, by minimizing
a loss function over a training set $L(\thetab)$, \ie, $\thetab^{*} = \argmin_{\thetab} L(\thetab)$.
Then, given an \emph{unseen} feature vector $\mathbf{x}_i$ from the test set, the classifier predicts $f_{\thetab}(\mathbf{x}_i) = 1$ if $d_{\thetab^{*}}(\mathbf{x}_i) \geq 0$ and $f_{\thetab}(\mathbf{x}_i) = -1$
other\-wise, where $d_{\thetab^{*}}(\mathbf{x})$ denotes the signed distance from the feature vector $\mathbf{x}$ to the decision boundary.

If class labels in the training set are correlated with one or more sensitive attributes $\{\mathbf{z}_i\}_{i=1}^{N}$ (\eg, gender, race), the percentage of users with a certain sensitive attribute having $d_{\thetab^{*}}(\mathbf{x}_i) \geq 0$ may differ dramatically from the
percentage of users without this sensitive attribute value having $d_{\thetab^{*}}(\mathbf{x}_i) \geq 0$ (\ie, the classifier may suffer from disparate impact).
Note that this may happen even if sensitive attributes are not used to construct the decision boun\-da\-ry but are correlated with one or more of user features, through indirect discrimination~\citep{pedreschi_discrimination}.

\vspace{-3mm}
\subsection{Fairness Definition} \label{sec:definition}
\vspace{-3.5mm}
First, to comply with \textbf{disparate treatment} criterion we specify that sensitive attributes are not used in decision making, \ie, $\{\mathbf{x}_i\}_{i=1}^{N}$ and $\{\mathbf{z}_i\}_{i=1}^{N}$ consist of disjoint feature sets.

Next, as discussed in Section~\ref{sec:intro}, our definition of \textbf{disparate impact} leverages the ``80\%-rule"~\citep{2005adverse}.
A decision boundary satisfies the ``80\%-rule" (or more generally the ``$p$\%-rule"), if the ratio between the percentage of users with a particular sensitive attribute value having $d_{\thetab}(\mathbf{x})\geq0$
and the percentage of users without that value having $d_{\thetab}(\mathbf{x})\geq0$ is no less than 80:100 ($p$:100). For a given  binary sensitive attribute $z \in \{0, 1\}$, one can write the $p$\%-rule as:
\begin{equation} \label{eq:rule}
\resizebox{0.95\columnwidth}{!}{$
\text{\scriptsize min}\left( \frac{P(d_{\thetab}(\mathbf{x}) \geq 0 | z=1)}{P(d_{\thetab}(\mathbf{x}) \geq 0 | z=0)} , \frac{P(d_{\thetab}(\mathbf{x}) \geq 0 | z=0)}{P(d_{\thetab}(\mathbf{x}) \geq 0 | z=1)} \right) \geq\frac{p}{100}.
$}
\end{equation}

Unfortunately, it is very challenging to directly incorporate the $p$\%-rule in the formulation of convex margin-based classifiers, since it is a non-convex function of the classifier parameters $\thetab$ and, therefore, it would lead to non-convex formulations, which are difficult to solve efficiently.
Secondly, as long as the user feature vectors lie on the same side of the decision boundary, the $p$\%-rule is invariant to changes in the decision boundary.
In other words, the $p$\%-rule is a function having saddle points. The presence of saddle points furthers complicate the procedure for solving non-convex optimization problems
 ~\citep{Dauphin_2014}.
To overcome these challenges, we next introduce a novel measure of decision boundary (un)fairness
which can be used as a proxy to efficiently design classifiers satisfying a given $p$\%-rule.

\vspace{-4mm}
\section {OUR APPROACH}\label{methodology}
\vspace{-4mm}

In this section, we first introduce our measure of decision boundary (un)fairness, the decision boundary covariance.
We  then derive two complementary formulations. The first formulation ensures compliance with disparate impact doctrine in its basic form (ensure a given $p$\%-rule) by maximizing accuracy subject to fairness constraints.
The second formulation guarantees fulfilling disparate impact's ``business necessity'' clause  by maximizing fairness subject to accuracy constraints.

For conciseness, we append a constant  1 to all feature vectors ($\mathbf{x_i}$) so that the linear classifier decision boundary equation $\thetab^{T}\mathbf{x} + b = 0$ reduces to $\thetab^{T}\mathbf{x} = 0$.

\vspace{-3mm}
\subsection{Decision Boundary Covariance} \label{cross-covariance}
\vspace{-3.5mm}
Our measure of decision boundary (un)fairness is defined as the covariance between the users'{} sensitive attributes, $\{ \mathbf{z}_i \}_{i=1}^{N}$, and the
signed distance from the users'{} feature vectors to the decision boundary, $\{ d_{\thetab}(\mathbf{x}_i) \}_{i=1}^{N}$, \ie:
\begin{eqnarray} \label{eq:fairness-definition}
\vspace{-2mm}
\nonumber
\cov(\mathbf{z}, d_{\thetab}(\mathbf{x})) &=&  \EE[ (\mathbf{z} - \bar{\mathbf{z}}) d_{\thetab}(\mathbf{x}) ] - \mathbb{E}[ (\mathbf{z} - \bar{\mathbf{z}}) ]  \bar{d}_{\thetab}(\mathbf{x})  \\
& \approx& \frac{1}{N} \sum_{i=1}^{N} \left(\mathbf{z}_i - \bar{\mathbf{z}}\right) d_{\thetab}(\mathbf{x}_i) ,
\vspace{-5mm}
\end{eqnarray}
where $\mathbb{E}[ (\mathbf{z} - \bar{\mathbf{z}}) ]  \bar{d}_{\thetab}(\mathbf{x})$ cancels out since $\mathbb{E}[ (\mathbf{z} - \bar{\mathbf{z}}) ]  = 0$. Since in linear models for classification, such as logistic regression
or linear SVMs, the decision boundary is simply the hyperplane defined by $\thetab^{T} \mathbf{x} = 0$, Eq.~\eqref{eq:fairness-definition} reduces to $\frac{1}{N} \sum_{i=1}^{N} \left(\mathbf{z}_i - \bar{\mathbf{z}}\right) \thetab^{T} \mathbf{x}_i$.

In contrast with the $p$\%-rule (Eq.~\ref{eq:rule}),
the decision boundary covariance (Eq.~\ref{eq:fairness-definition}) is a convex function with respect to the decision boundary parameters $\thetab$, since $d_{\thetab}(\mathbf{x}_i)$ is convex with respect to $\thetab$ for all linear, convex margin-based classifiers.
\footnote{For non-linear convex margin-based classifiers like non-linear SVM, equivalent of $d_{\thetab}(\mathbf{x}_i)$ is still convex in the transformed kernel space. See Appendix~\ref{sec:svm_formulation} for details.}
 Hence,
it can be easily included in the formulation of these classifiers
 without increasing the complexity of their training.

Moreover, note that, if a decision boundary satisfies the ``100\%-rule",
\ie,
\begin{equation} \label{eq:prob_equal}
P(d_{\thetab}(\mathbf{x}) \geq 0 | z = 0) = P(d_{\thetab}(\mathbf{x}) \geq 0 | z = 1),
\end{equation}
then the (empirical) covariance will be approximately zero for a sufficiently large training set.

\vspace{-3mm}
\subsection{Maximizing Accuracy Under Fairness Constraints} \label{constraints}
\vspace{-3.5mm}

In this section, we design classifiers that maximize accu\-ra\-cy subject to fairness constraints (\eg, a specific $p$\%-rule),
and thus may be used to ensure compliance with the disparate impact doctrine in its basic form.

To this end, we find the decision boundary parameters $\thetab$ by minimizing the corres\-pon\-ding loss function over the training set under fairness constraints, \ie:
\begin{equation} \label{eq:general-classifier-no-discrimination}
	\begin{array}{ll}
		\mbox{minimize} & L(\thetab) \\
		\mbox{subject to} & \frac{1}{N} \sum_{i=1}^{N} \left(\mathbf{z}_i - \bar{\mathbf{z}}\right) d_{\thetab}(\mathbf{x}_i) \leq \mathbf{c}, \\
		& \frac{1}{N} \sum_{i=1}^{N} \left(\mathbf{z}_i - \bar{\mathbf{z}}\right)d_{\thetab}(\mathbf{x}_i) \geq -\mathbf{c},
	\end{array}
\end{equation}
where $\mathbf{c}$ is the covariance threshold, which specifies an upper bound on the covariance between each sensitive attribute and the signed distance from the feature vectors to the
decision boundary.
In this formulation,  $\mathbf{c}$ trades off fairness and accuracy,  such that as we decrease $\mathbf{c}$ towards zero, the resulting classifier will satisfy a larger $p$\%-rule but will potentially suffer from a larger loss in accuracy.
Note that since the above optimization problem is convex, our scheme ensures that the trade-off between the classifier loss function and decision boundary covariance is Pareto optimal.

\xhdr{Remarks}
It is important to note that the distance to the margin, $d_{\thetab}(\mathbf{x})$, only depends on the non-sensitive features $\mathbf{x}$ and, therefore, the sensitive features $\mathbf{z}$ are not needed while making decisions.  In other words, we account for
\textit{disparate treatment},
by removing the sensitive features from the decision making process and, for
\textit{disparate impact},
by adding fairness constraints during the training process of the classifier.
Additionally, the constrained optimization problem~\eqref{eq:general-classifier-no-discrimination}, can also be written as a regularized optimization problem by making use of its dual form, in which the fairness constraints are moved to the objective and the corresponding Lagrange multipliers act as regularizers.

Next, we specialize problem~\eqref{eq:general-classifier-no-discrimination} for logistic regression classifiers.

\xhdr{Logistic Regression}
In logistic regression classifiers, one maps the feature vectors $\mathbf{x}_i$ to the class labels $y_i$ by means of a probability distribution:
\begin{equation}
p(y_i = 1 | \mathbf{x}_i, \thetab) = \frac{1}{1 + e^{-\thetab^{T}  \mathbf{x}_i} },
\end{equation}
where  $\thetab$ is obtained by solving a ma\-xi\-mum likelihood problem over the training set, \ie, $\thetab^{*} = \argmin_{\thetab} - \sum_{i =1}^N \log p(y_i | \mathbf{x}_i, \thetab)$.
Thus, the corresponding loss function is given by $- \sum_{i =1}^N \log p(y_i | \mathbf{x}_i, \thetab)$, and problem~\eqref{eq:general-classifier-no-discrimination} adopts the form:
\begin{align}
       \begin{array}{ll} \nonumber
        		\mbox{minimize} & - \sum_{i=1}^{N} \log p(y_i | \mathbf{x}_i, \thetab) \hspace{11.5mm} \quad \\
    \end{array} &
    \\
     \begin{array}{ll}
      \mbox{subject to}  & \frac{1}{N} \sum_{i=1}^{N} \left(\mathbf{z}_i - \bar{\mathbf{z}}\right)\thetab^{T}  \mathbf{x}_i \leq \mathbf{c},   \\
       &  \frac{1}{N} \sum_{i=1}^{N} \left(\mathbf{z}_i - \bar{\mathbf{z}}\right) \thetab^{T}  \mathbf{x}_i \geq -\mathbf{c},
    \end{array}& \label{eq:logistic-regression-ml-no-discrimination}\end{align}

Appendix~\ref{sec:svm_formulation} presents the specialization of our formulation for both linear and non-linear SVM classifiers.

\vspace{-3mm}
\subsection{Maximizing Fairness Under Accuracy Constraints}\label{classification_accuracy_bound}
\vspace{-3.5mm}

In the previous section, we designed classifiers that maximize accuracy subject to fairness constraints. However, if the underlying correlation between the class labels and the sensitive attributes in the training set is very high, enforcing fairness constraints may results in underwhelming performance (accuracy) and thus be unacceptable in terms of business objectives.
Disparate impact's ``business necessity'' clause accounts for such scenarios by allowing \textit{some} degree of disparate impact in order to meet performance constraints. However, the employer needs to ensure that the decision making causes \textit{least possible} disparate impact under the given performance (accuracy) constraints~\citep{barocas_2016}.
To accommodate such scenarios, we now propose an alternative formulation that maximizes fairness (minimizes disparate impact) subject to accuracy constraints.

\begin{figure*}[ht]

	\centering
	\subfloat[Maximizing  accuracy under fairness constraints]{
	\begin{tabular}{cc}
	\includegraphics[width=.22\textwidth, trim = 2.2cm 1.6cm 2cm 2.5cm]{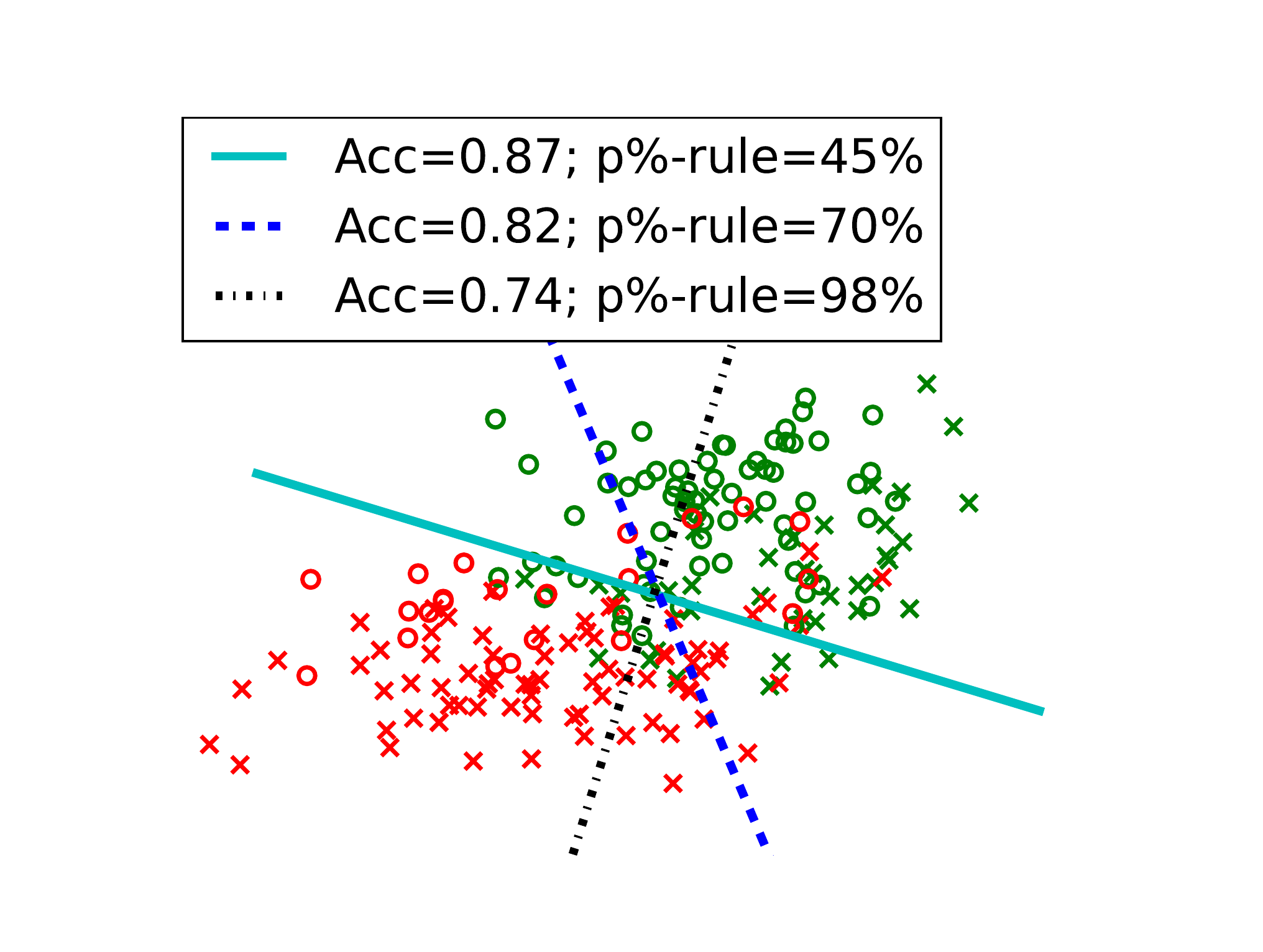} &
	\includegraphics[width=.22\textwidth, trim = 2.2cm 1.6cm 2cm 2.5cm]{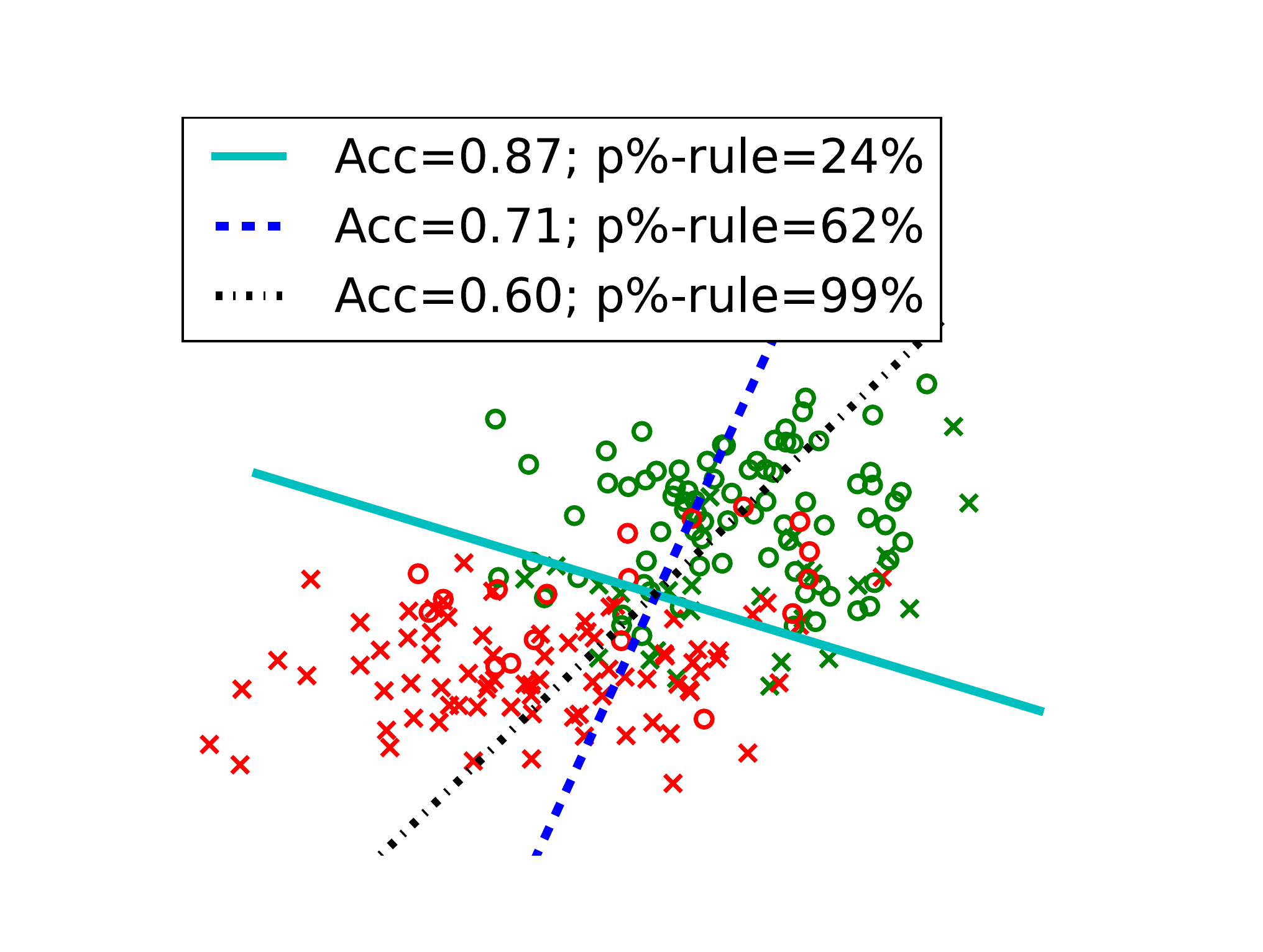}\\
	$\phi = \pi/4$ & $\phi = \pi/8$
	\end{tabular} \label{fig:decision-boundary-synthetic} 	}
	\subfloat[Maximizing fairness under accuracy constraints]{
	\begin{tabular}{cc}
         \includegraphics[width=.22\textwidth, trim = 2.2cm 1.6cm 2cm 2.5cm]{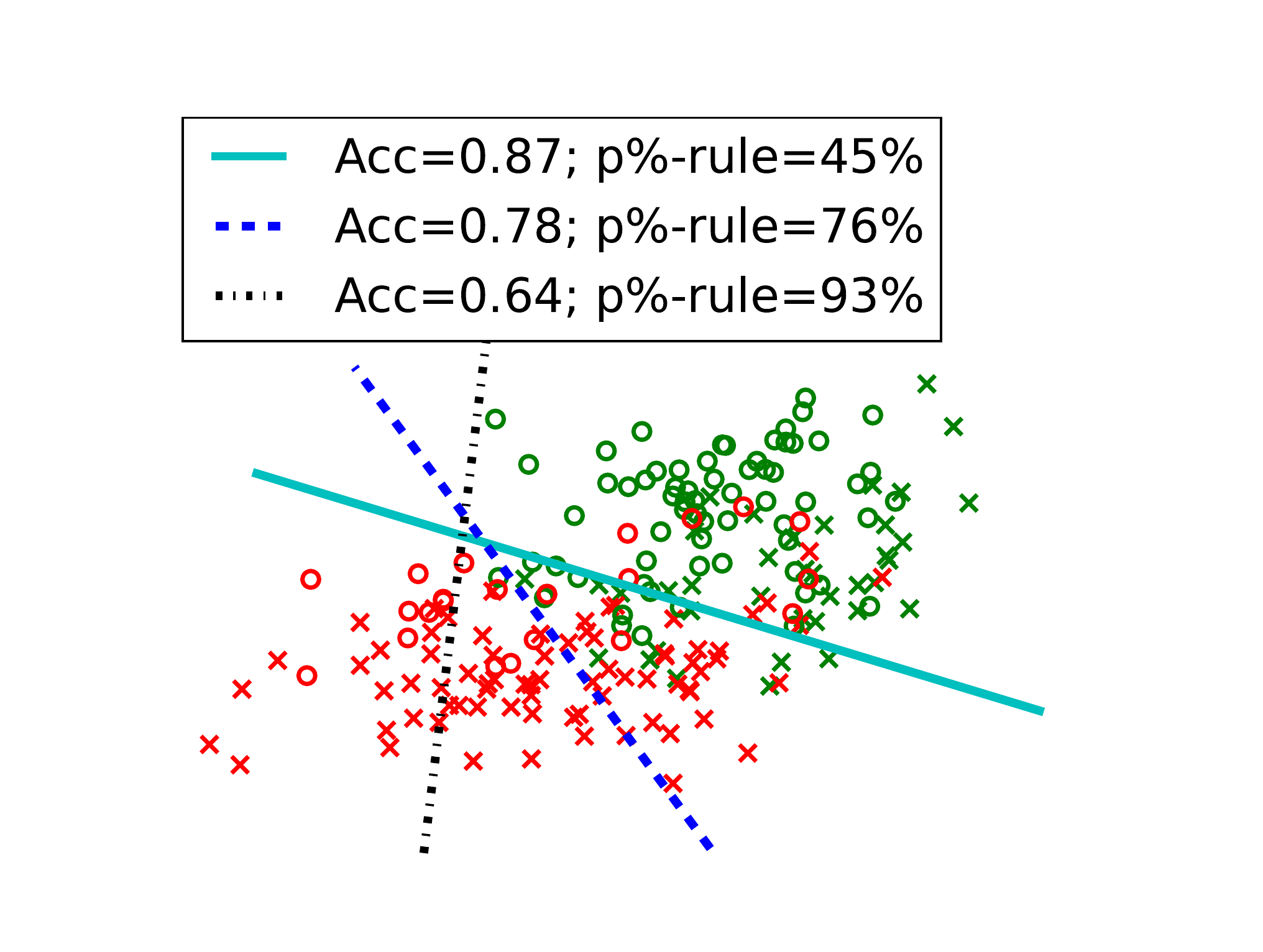} &
	 \includegraphics[width=.22\textwidth, trim = 2.2cm 1.6cm 2cm 2.5cm]{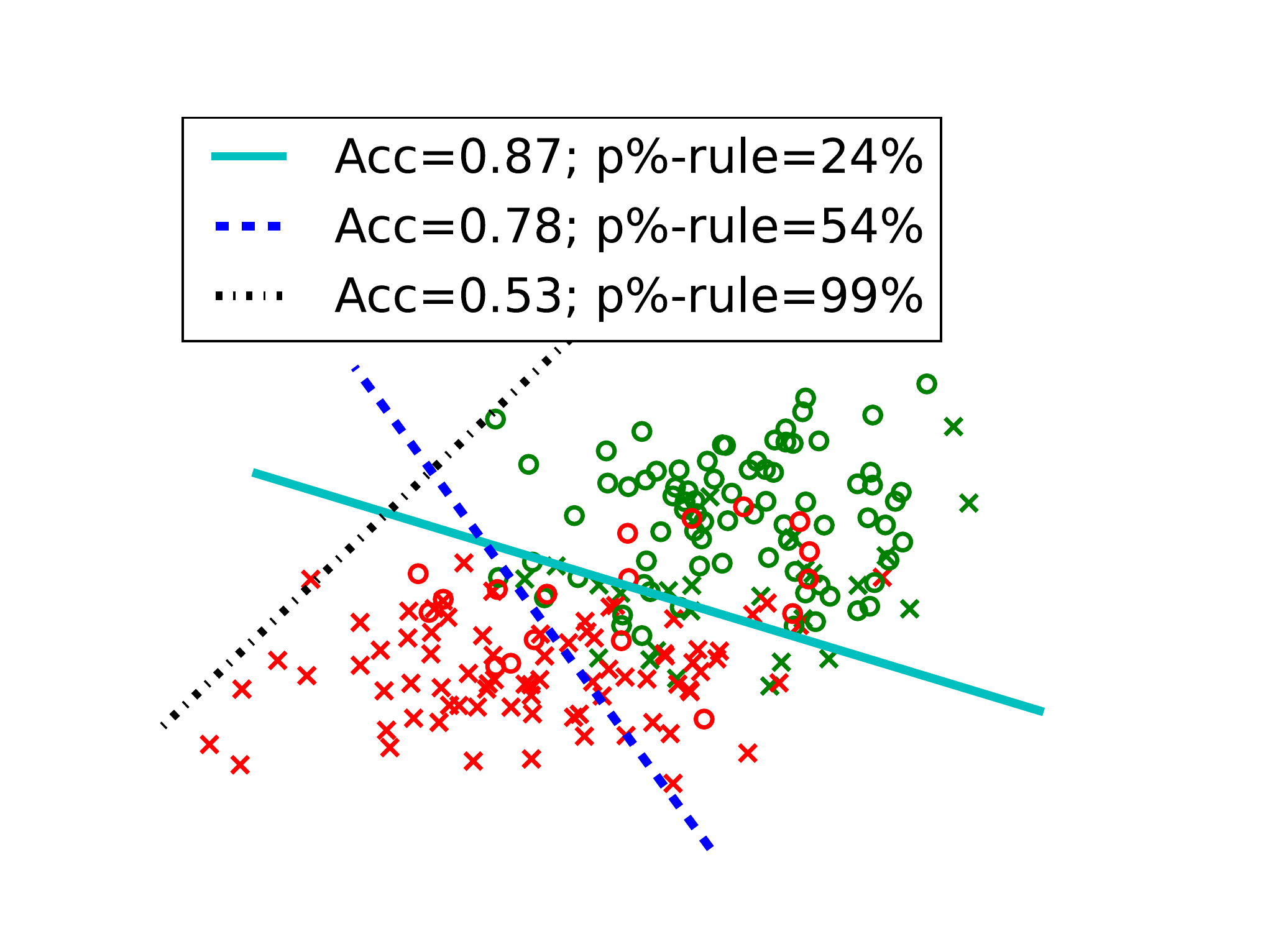} \\
	 $\phi = \pi/4$ & $\phi = \pi/8$
	 \end{tabular}  \label{fig:decision-boundary-synthetic-no-misclass}}
	\vspace{-3mm}
	\caption{The solid light blue lines show the decision boundaries for logistic regressors without fairness constraints. The dashed lines show the decision boundaries for fair logistic regressors trained (a) to maximize accuracy under fairness constraints and (b) to maximize fairness under fine-grained accuracy constraints, which prevents users with $z=1$ (circles) labeled as positive by the unconstrained classifier from being moved to the negative class. Each column corresponds to a dataset, with different correlation value between sensitive attribute values (crosses vs circles) and class labels (red vs green).}
 \vspace{-3mm}
\end{figure*}

To this aim, we find the decision boundary parameters $\thetab$ by minimizing the corresponding (absolute) decision boundary covariance over the training set under constraints on the classifier loss function, \ie:
 \begin{equation} \label{eq:ml-no-discrimination-accuracy-bound}
	\begin{array}{ll}
		\mbox{minimize} & |\frac{1}{N} \sum_{i=1}^{N} \left(\mathbf{z}_i - \bar{\mathbf{z}}\right) d_{\thetab}( \mathbf{x}_i) | \\
		\mbox{subject to} & L(\thetab) \leq (1+\gamma) L(\thetab^{*}),
	\end{array}
\end{equation}
where $L(\thetab^{*})$ denotes the optimal loss over the training set provided by the unconstrained classifier and $\gamma \geq 0$ specifies the maximum additional loss with respect to the loss provided by the unconstrained
classifier. Here, we can ensure maximum fairness with no loss in accuracy by setting $\gamma = 0$.
As in Section~\ref{constraints}, it is possible to specialize problem~\eqref{eq:ml-no-discrimination-accuracy-bound} for the same classifiers and show that the formulation remains convex.

\xhdr{Fine-Grained Accuracy Constraints}
In many classifiers, including logistic regression and SVMs, the loss function (or the dual of the loss function) is addi\-tive over the points in the training set, \ie, $L(\thetab)= \sum_{i=1}^{N} L_i(\thetab)$, where $L_i(\thetab)$ is the individual
loss associated with the $i$-th point in the training set.
Moreover, the individual loss $L_i(\thetab)$ typically tells us how \emph{close} the predicted label $f(\mathbf{x}_i)$ is to the true label $y_i$, by means of the signed distance to the decision boundary.
Therefore, one may think of incorporating loss constraints for a certain set of users, and consequently, prevent individual users originally classified as positive (by the unconstrained classifier) from being classified as negative by the constrained classifier.
To do so, we find the decision boundary parameters $\thetab$ as:
\begin{equation} \label{eq:ml-no-discrimination-accuracy-set}
	\begin{array}{ll}
		\mbox{minimize} & |\frac{1}{N} \sum_{i=1}^{N} \left(\mathbf{z}_i - \bar{\mathbf{z}}\right) \thetab^{T} \mathbf{x}_i | \\
		\mbox{subject to} & L_i(\thetab) \leq (1+\gamma_i) L_i(\thetab^{*})  \quad \forall i \in \{1, \ldots, N\},
	\end{array}
\end{equation}
where $L_i(\thetab^{*})$ is the individual loss associated to the $i$-th user in the training set provided by the unconstrained classifier and $\gamma_i \geq 0$ is her allowed additional loss.
For example, in the case of logistic regression classifier,  $\thetab^{*}= \argmin_{\thetab} \sum_{i=1}^{N} -\log p(y_i | \mathbf{x}_i, \thetab)$ and
the losses for individual points are  $ L_i(\thetab) = - \log p(y_i | \mathbf{x}_i, \thetab)$.
 Now, if we set $\gamma_i=0$, we are enforcing that the probability of the $i$-th user to be mapped in the positive class to be
equal or higher than in the original (unconstrained) classifier.

\vspace{-3mm}
\section {EVALUATION}\label{evaluation}
\vspace{-3mm}
We evaluate our framework on several synthetic and real-world datasets.
We first experiment with our first formulation and show that it allows for fine-grained fairness control,
 often at a minimal loss in
accuracy.
Then, we validate our second formulation, which allows maximizing fairness under accuracy constraints, and also provides guarantees on avoiding negative classification of certain individual users or group of users.

Here, we adopt the $p$\%-rule~\citep{2005adverse} as our true measure of fairness.
However, as shown in Appendix~\ref{app:real_exp}, we obtain similar results if we consider another measure of fairness used by some of the previous studies in this area.

\vspace{-3mm}
\subsection{Experiments on Synthetic Data} \label{sec:synthetic}
\vspace{-3.5mm}
\xhdr{Fairness constraints vs accuracy constraints}
To simulate different degrees of disparate impact in classification outcomes, we generate two synthetic datasets with diffe\-rent levels of correlation between a single, binary sensitive attribute and class labels. We then train two types of logistic re\-gression
classifiers: one type maximizes accuracy subject to fairness constraints (Section~\ref{constraints}), and the other maximizes fairness under fine-grained accuracy
constraints (Section~\ref{classification_accuracy_bound}).

Specifically,
we ge\-ne\-rate $4$,$000$ binary class labels uniformly at random and assign a 2-dimensional user feature vector per label by drawing samples
from two different Gaussian distributions:
$p(x|y$$=$$1) = N([2; 2], [5,\ 1; 1,\  5])$
and $p(x|y$$=$$-1) = N([-2; -2], [10,\ 1; 1,\ 3])$.
Then, we draw each user'{}s sensitive attri\-bute $z$ from a Bernoulli distribution: $p(z=1) = p(\mathbf{x}'|y=1) / (p(\mathbf{x}'|y=1) + p(\mathbf{x}'|y=-1))$,
where $\mathbf{x}' = [\cos(\phi),\ -\sin(\phi);  \sin(\phi),\ \cos(\phi)] \mathbf{x}$ is simply a rotated version of the feature vector, $\mathbf{x}$.

We generate datasets with two values for the parameter $\phi$, which controls the correlation between the sensitive attribute and the class labels (and hence, the resulting disparate impact). Here, the closer $\phi$ is to zero, the higher the correlation.
Finally, we trained both types of constrained classifiers on each dataset.

Fig.~\ref{fig:decision-boundary-synthetic} shows the decision boundaries provided by the classifiers that maximize accuracy under fairness constraints for two different correlation
values
and two (successively decreasing) covariance thresholds, $\mathbf{c}$. We compare these boundaries against the unconstrained decision boundary  (solid line).
As expected, given the data generation process, fairness constraints map into a rotation of the decision boundary (dashed lines), which is greater
as we decrease threshold value $\mathbf{c}$ or increase the correlation in the original data (from $\phi = \pi/4$ to $\phi = \pi/8$).
This movement of the decision boundaries shows that our fairness constraints are successfully undoing (albeit in a highly controlled setting) the rotations we used to induce disparate impact in the dataset. Moreover, a smaller covariance threshold (a larger rotation) leads to a more fair solution, although, it comes at a larger cost in accuracy.

Fig.~\ref{fig:decision-boundary-synthetic-no-misclass} shows the decision boundaries provided by the classifiers that maximize fairness under fine-grained accuracy
constraints.
Here, the fine-grained accuracy constraints ensure that the
users with $z=1$
classified as positive by the unconstrained classifier (circles above
 the solid line) are not labeled as negative by the fair classifier.
The decision boundaries provided by this formulation, in contrast to the previous one, are rotated \textit{and shifted} versions of the unconstrained boundary.
Such shifts enable the constrained classifiers to avoid negatively classifying users specified in the constraints.

We also illustrate how the decision boundary of a non-linear classifier, a SVM with RBF kernel, changes under fairness constraints in Appendix~\ref{app:syn_exp}.

\begin{figure*}[t]
\vspace{-10mm}
	\centering
	\subfloat[Loss vs. Cov.]{
	\label{fig:pareto}
	\begin{tabular}{c}
	\multicolumn{1}{c}{\hspace{4mm} \includegraphics[width=.038\textwidth, angle=-90]{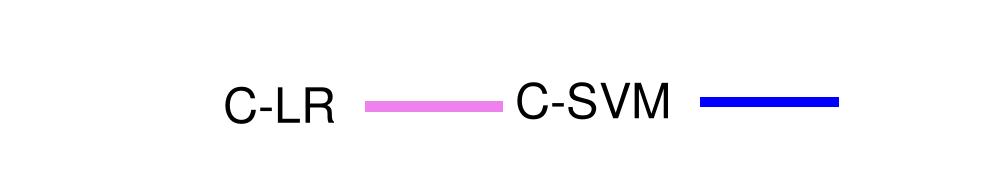}} \\ [-3mm]
	\includegraphics[width=.15\textwidth, angle=-90, trim = 0.2cm 0.5cm 0.2cm 0.5cm]{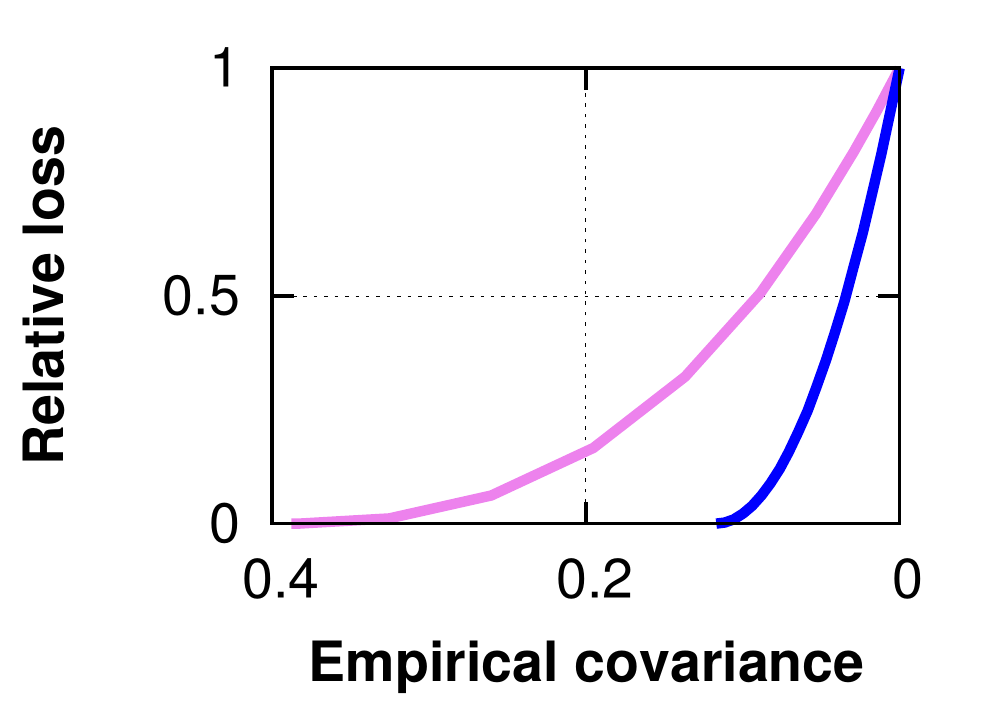}\\ [-3mm]
	\multicolumn{1}{c}{\hspace{4mm} \includegraphics[width=.038\textwidth, angle=-90]{legend_cov_vs_p_rule}} \\[-3mm]
    \setcounter{subfigure}{1}\includegraphics[width=.15\textwidth, angle=-90, trim = 0.2cm 0.5cm 0.2cm 0.5cm]{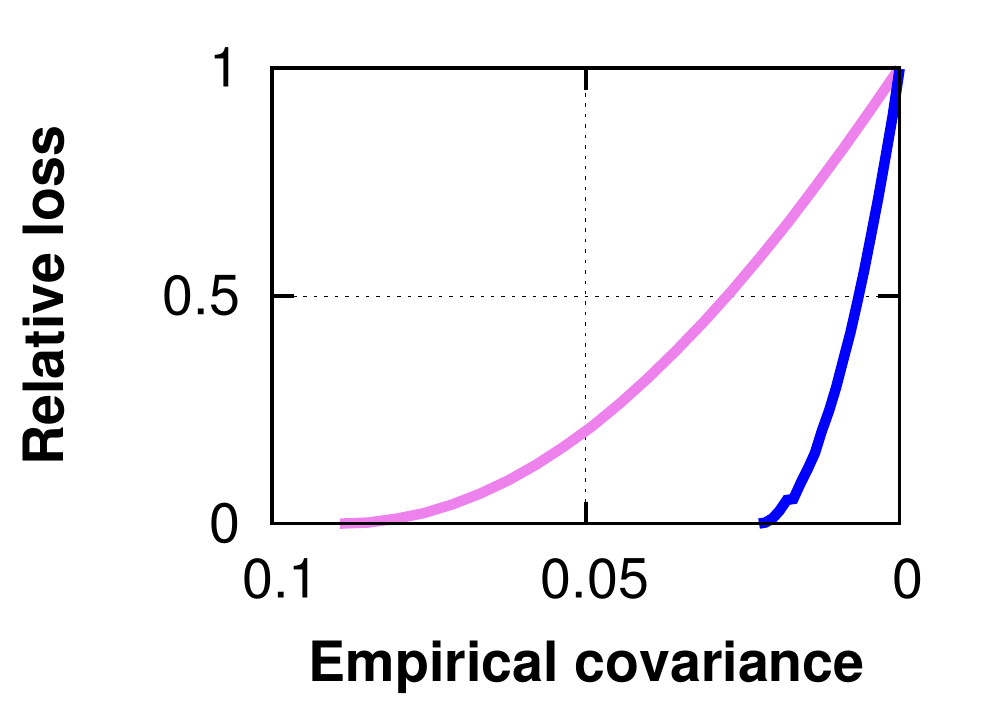}  \\ \\ [-3mm]
	\small{ \hspace{2mm} Adult and Bank}
	\end{tabular}
	}
	\hspace{-6mm}
	\subfloat[Cov. vs. $p$\%-rule]{
	\label{fig:p-rule_vs_cross-cov}
	\begin{tabular}{c}
	\multicolumn{1}{c}{\hspace{4mm} \includegraphics[width=.035\textwidth, angle=-90, trim = 0cm 0.5cm 0.2cm 0.5cm]{legend_cov_vs_p_rule}} \\ [-3mm]
	\includegraphics[width=.15\textwidth, angle=-90, trim = 0.2cm 0.5cm 0.2cm 0.5cm]{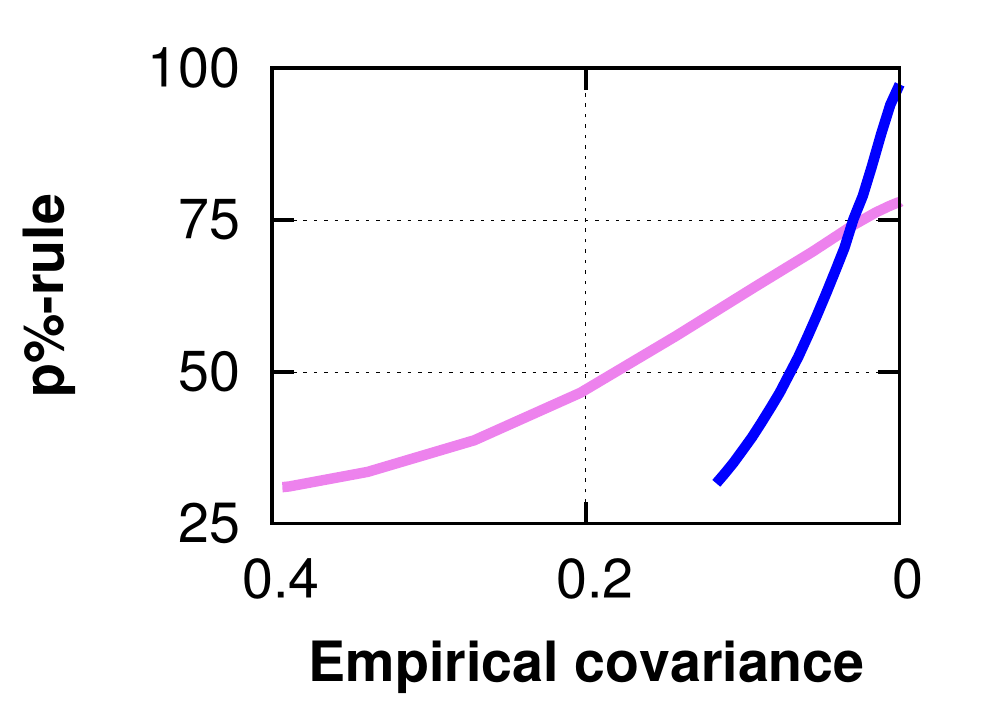}\\ [-3mm]
	\multicolumn{1}{c}{\hspace{4mm} \includegraphics[width=.035\textwidth, angle=-90, trim = 0cm 0.5cm 0.2cm 0.5cm]{legend_cov_vs_p_rule}} \\[-3mm]
    \includegraphics[width=.15\textwidth, angle=-90, trim = 0.2cm 0.5cm 0.2cm 0.5cm]{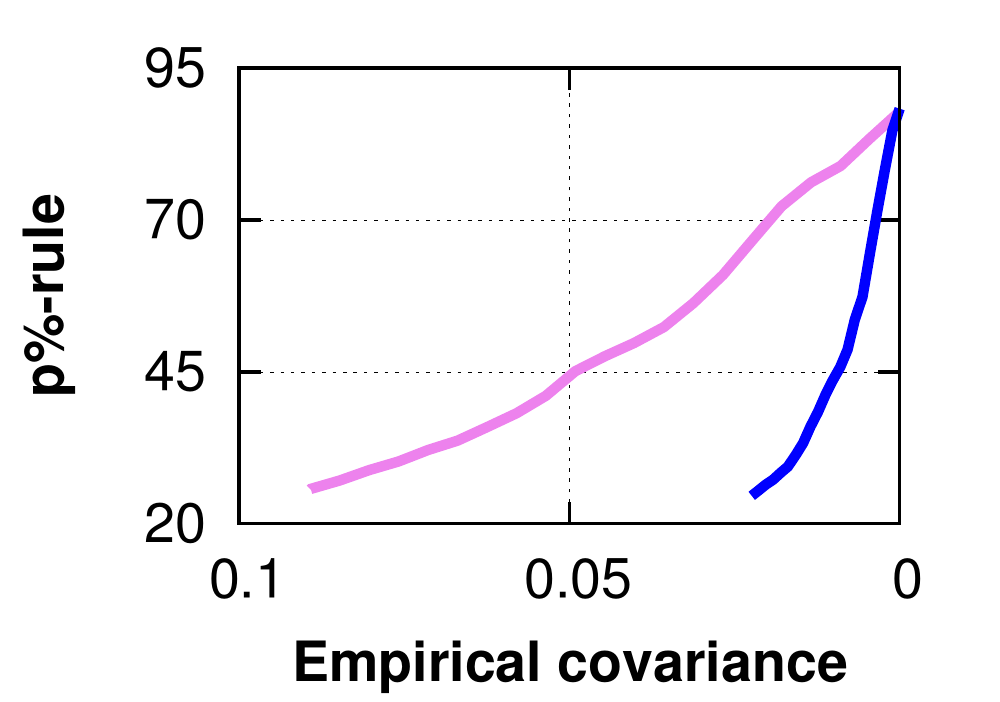}  \\ \\ [-3mm]
	\small{ \hspace{2mm} Adult and Bank}
	\end{tabular}
	}
	\subfloat[Single binary sensitive attribute]{
	\label{fig:accuracy-all}
	\begin{tabular}{cc}
	\multicolumn{2}{c}{\hspace{-3mm} \includegraphics[width=.035\textwidth, angle=-90, trim = 0cm 0.5cm 0.2cm 0.5cm]{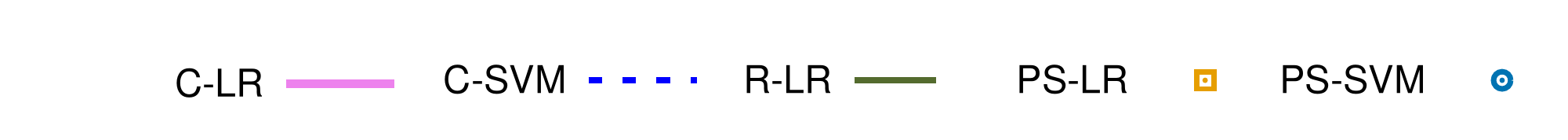}} \\ [-3mm]
	\hspace{-3mm}  \includegraphics[width=.15\textwidth, angle=-90, trim = 0.2cm 0.5cm 0.2cm 0.5cm]{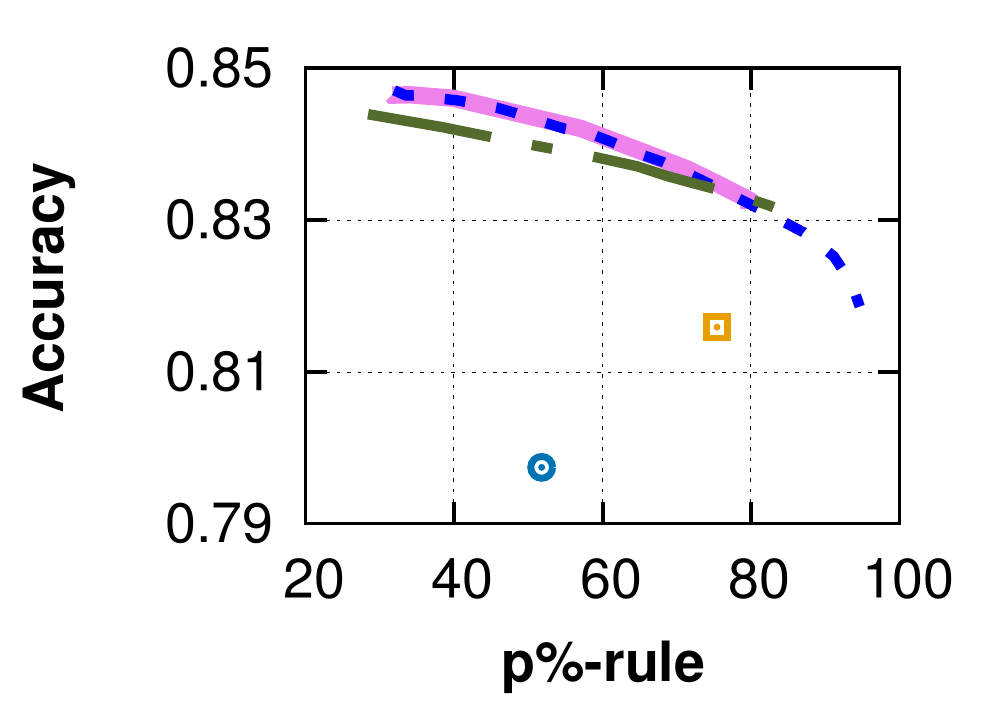}   &
	 \includegraphics[width=.15\textwidth, angle=-90, trim = 0.2cm 0.5cm 0.2cm 0.5cm]{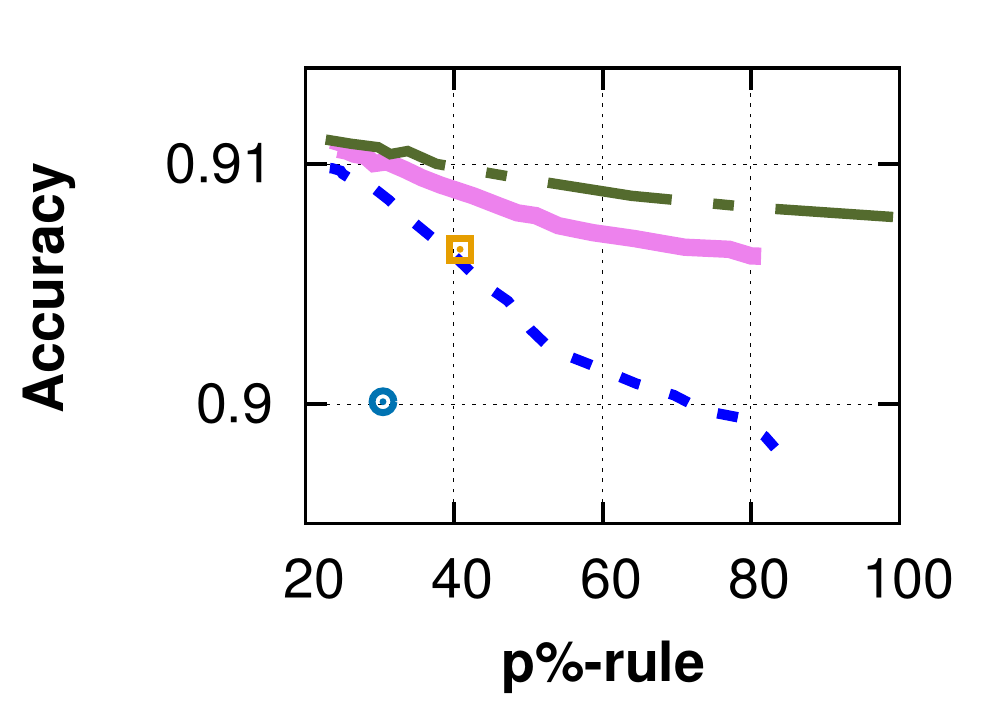} \\ [-3mm]
	\multicolumn{2}{c}{\hspace{-3mm} \includegraphics[width=.035\textwidth, angle=-90, trim = 0cm 0.5cm 0.2cm 0.5cm]{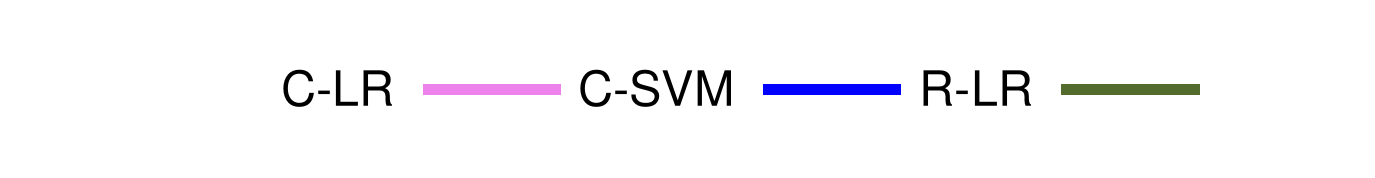}}  \\  [-3mm]
	\hspace{-3mm} \includegraphics[width=.15\textwidth, angle=-90, trim = 0.2cm 0.5cm 0.2cm 0.5cm]{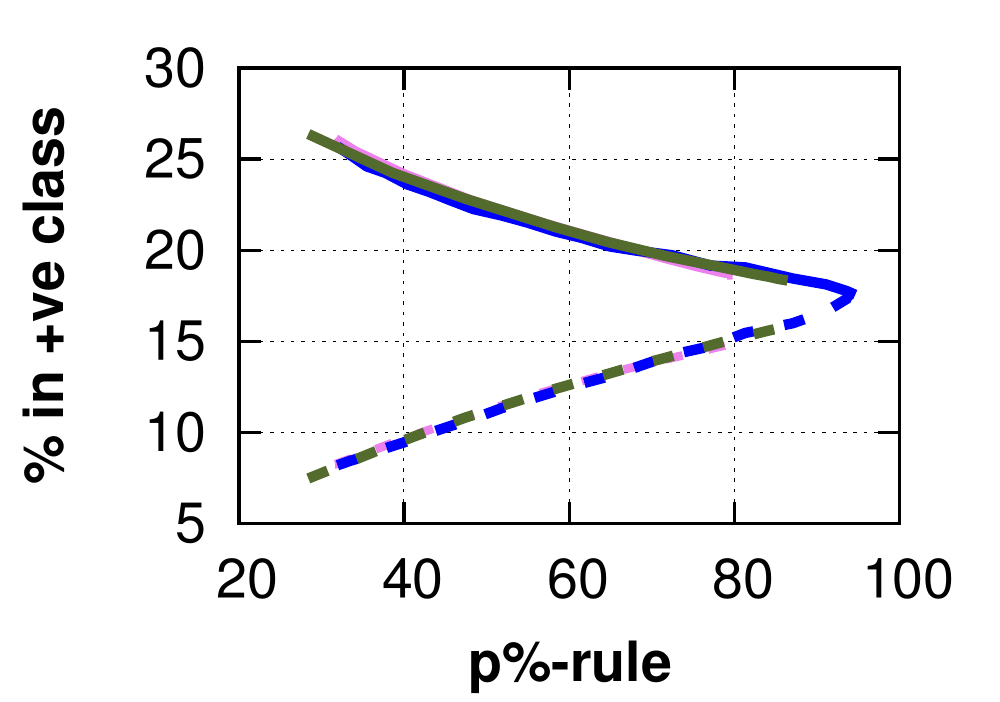}  &
	\includegraphics[width=.15\textwidth, angle=-90, trim = 0.2cm 0.5cm 0.2cm 0.5cm]{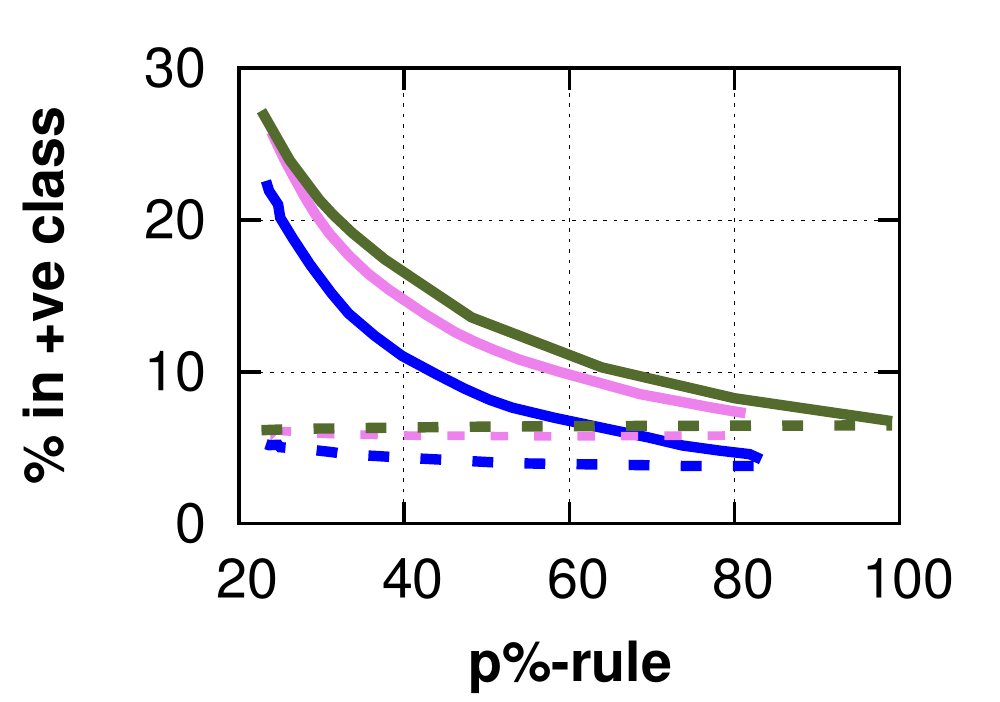} \\ \\ [-3mm]
	\small{\hspace{2mm}Adult} & \small{ \hspace{2mm}Bank}
	\end{tabular}
	}
	\vspace{-2mm}
	\caption{
	[Maximizing accuracy under fairness constraints: single, binary sensitive attribute]
		Panels in (a) show the trade-off between the empirical covariance in Eq.~\ref{eq:fairness-definition} and the relative loss (with respect to the unconstrained classifier), for the Adult (top) and Bank (bottom) datasets. Here each pair of (covariance, loss)
		values is guaranteed to be Pareto optimal by construction.
		Panels in (b) show the correspondence between the empirical covariance and the $p$\%-rule for classifiers trained under fairness constraints.
		Panels in (c) show the accuracy against $p$\%-rule value (top) and the percentage of protected (dashed) and non-protected (solid) users in the positive class against the $p$\%-rule value (bottom).
	}
	\vspace{-5mm}
\end{figure*}

\begin{figure}[!h]
\vspace{-4mm}
	\centering
	\subfloat{
	\begin{tabular}{cc}
	\multicolumn{2}{c}{\hspace{-3mm} \includegraphics[width=.038\textwidth, angle=-90]{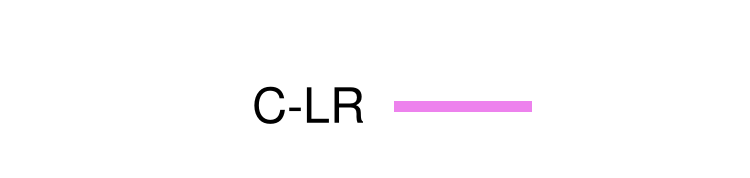}} \\ [-3mm]
	 \hspace{-3mm}  \includegraphics[width=.15\textwidth, angle=-90, trim = 0.2cm 0.5cm 0.2cm 0.5cm]{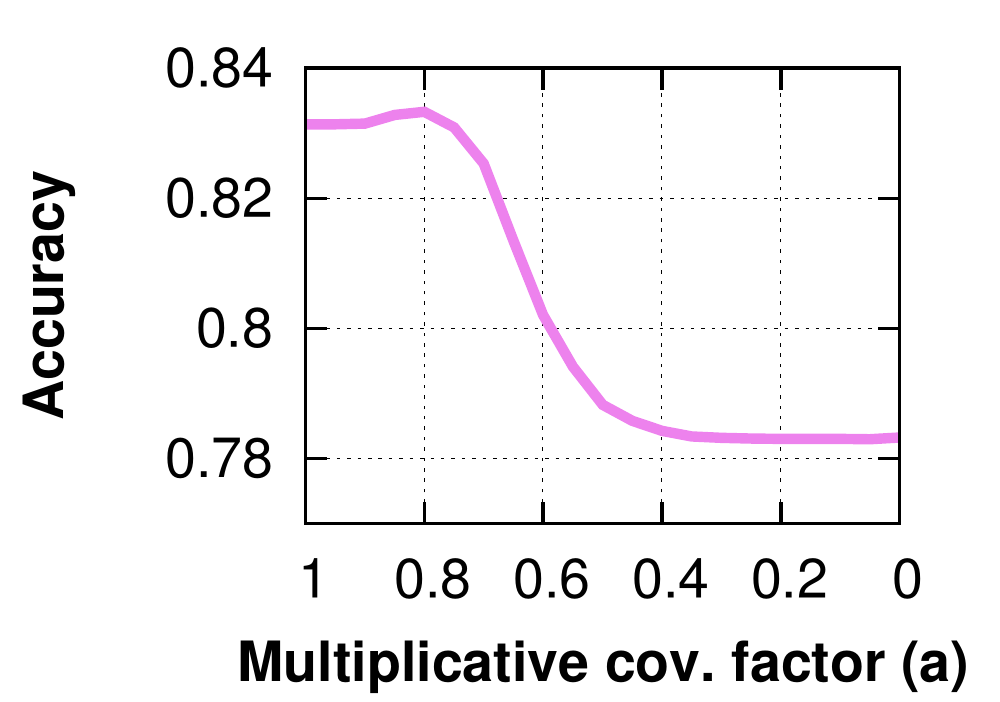}   &
	 \includegraphics[width=.15\textwidth, angle=-90, trim = 0.2cm 0.5cm 0.2cm 0.5cm]{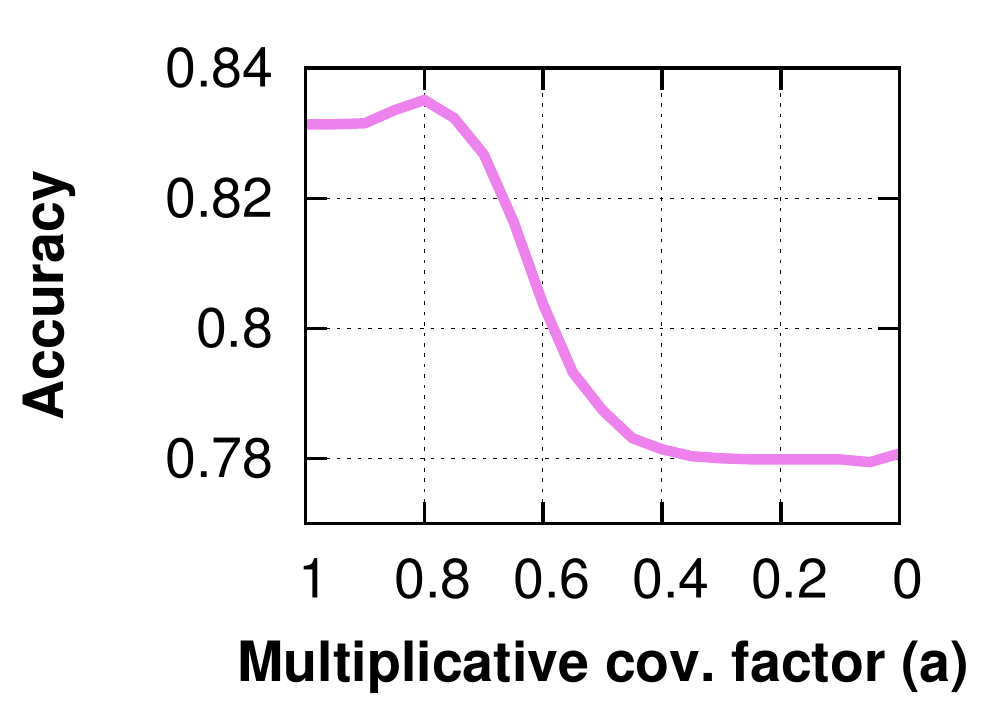} \\ [-3mm]
	\multicolumn{2}{c}{\hspace{-5mm} \includegraphics[width=.035\textwidth, angle=-90, trim = 0cm 0.5cm 0.3cm 0.5cm]{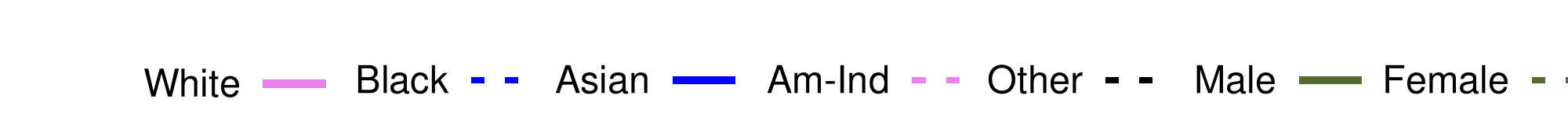}}  \\ [-3mm]
	\hspace{-3mm} \includegraphics[width=.15\textwidth, angle=-90, trim = 0.2cm 0.5cm 0.2cm 0.5cm]{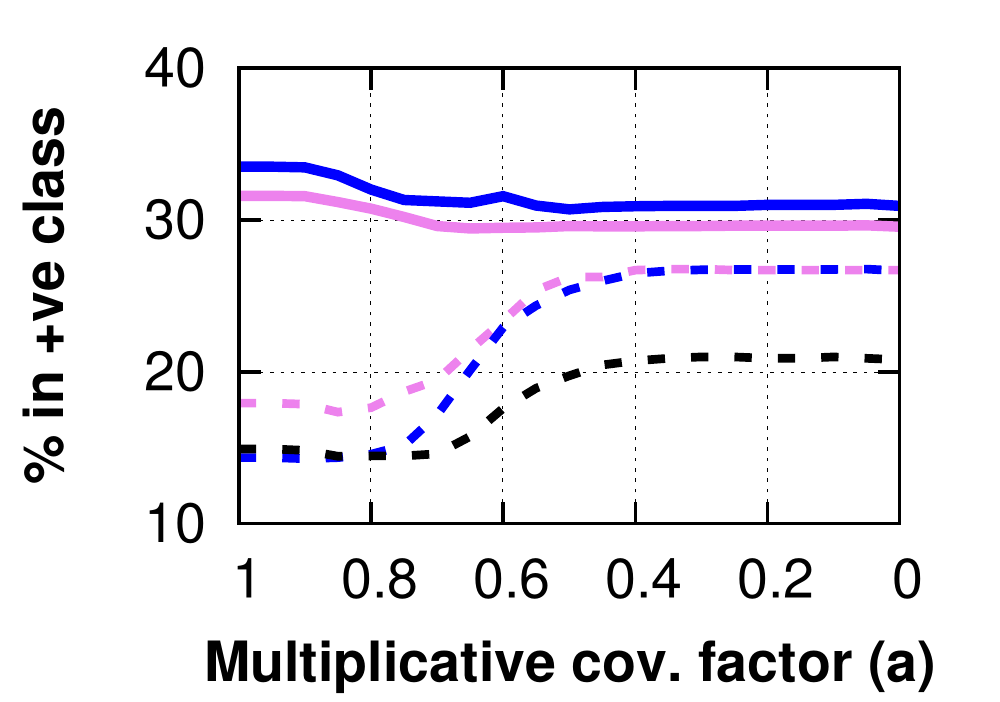}  &
	\includegraphics[width=.15\textwidth, angle=-90, trim = 0.2cm 0.5cm 0.2cm 0.5cm]{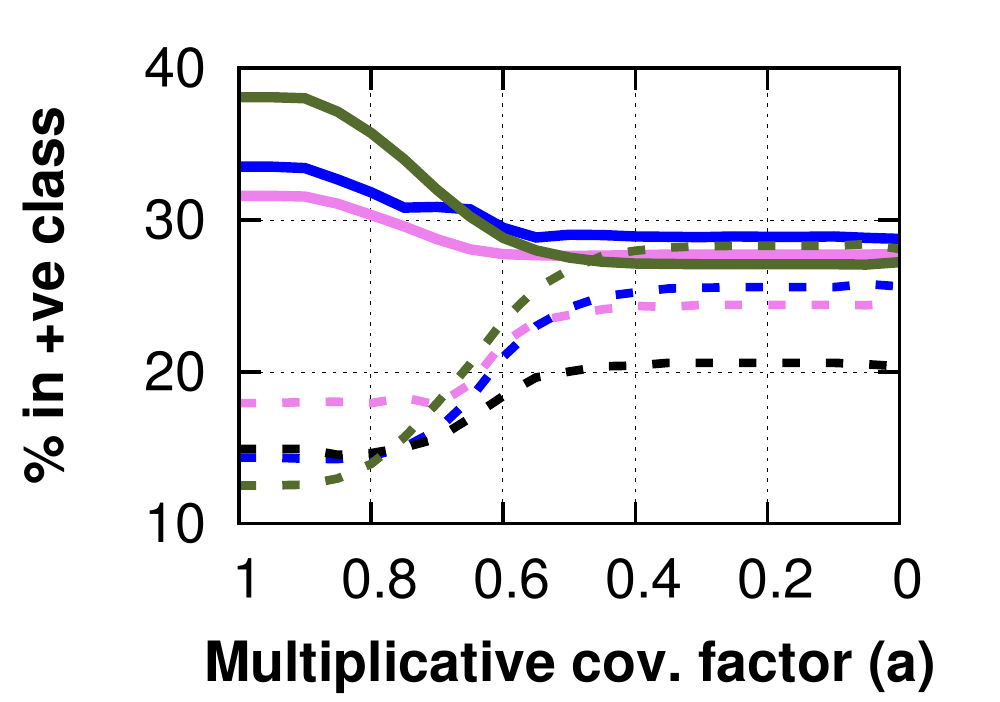}  \\ \\ [-3mm]

	\small{\hspace{2mm} Non-binary} & \small{ \hspace{2mm} Several}
	\end{tabular}
	}
	\vspace{-2mm}
	\caption{[Maximizing accuracy under fairness constraints: non-binary and several sensitive attributes]
	The figure shows accuracy (top) and percentage of users in positive class (bottom) against a multiplicative factor $a \in [0,1]$ such that $\mathbf{c} = a \mathbf{c}^{*}$, where $\mathbf{c}^{*}$ denotes the
	unconstrained classifier covariance.} \label{non_bin_several_first}
	\vspace{-4mm}
\end{figure}

\vspace{-3mm}
\subsection{Experiments on Real Data}\label{exp:real}
\vspace{-3.5mm}
\xhdr{Experimental Setup}
We experiment with two real-world datasets: The Adult income dataset~\citep{adult_dataset} and the Bank marketing dataset~\citep{bank_dataset}.
The Adult dataset contains a total of $45{,}222$ subjects, each with $14$ features (e.g., age, educational level)
and a binary label, which indicates whether a subject's incomes is above (positive class) or below (negative class) 50K USD. For
this dataset, we consider gender and race, respectively, as binary and non-binary (polyvalent) sensitive attributes.
The Bank dataset contains a total of $41{,}188$ subjects, each with 20 attributes  (e.g., marital status)
and a binary label, which indicates whether the client has subscribed (positive class) or not (negative class) to a term deposit. In this case, we consider age as (binary) sensitive attribute, which is discretized to indicate whether the client{}'s age is between 25 and 60 years.
For detailed statistics about the distribution of different sensitive attributes in positive class in these datasets, we refer the reader to Appendix~\ref{app:real_exp}.

For the sake of conciseness, while presenting the results for binary sensitive attributes, we refer to females and males, respectively, as protected and non-protected groups in Adult data. Similarly, in Bank data, we refer to users between age 25 and 60 as protected and rest of the users as non-protected group.
In our experiments, to obtain more reliable estimates of accuracy and fairness, we repeatedly split each dataset into a train (70\%) and test (30\%) set 5 times and report the average statistics for accuracy and fairness.

\begin{figure*}[t]
\vspace{-10mm}
	\centering
	\vspace{-4mm}
	\subfloat[Acc and $p$\%-rule]{
	\label{acc_and_prule_second}
	\begin{tabular}{cc} \\
	\multicolumn{2}{c}{\includegraphics[width=.035\textwidth, angle=-90, trim = 0cm 0.5cm 0.2cm 0.5cm]{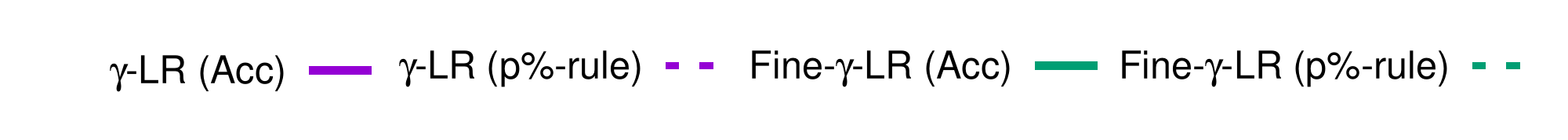}} \\ [-3mm]
	\includegraphics[width=.13\textwidth, angle=-90, trim = 0.2cm 0.5cm 0.2cm 0.5cm]{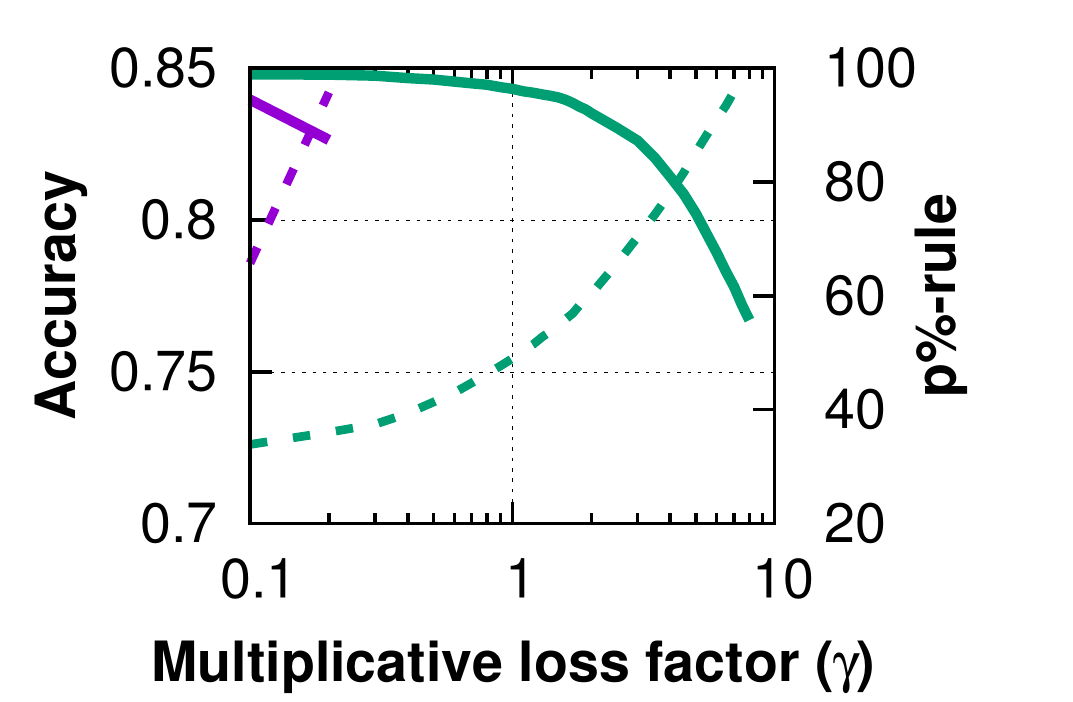} &
	\includegraphics[width=.13\textwidth, angle=-90, trim = 0.2cm 0.5cm 0.2cm 0.5cm]{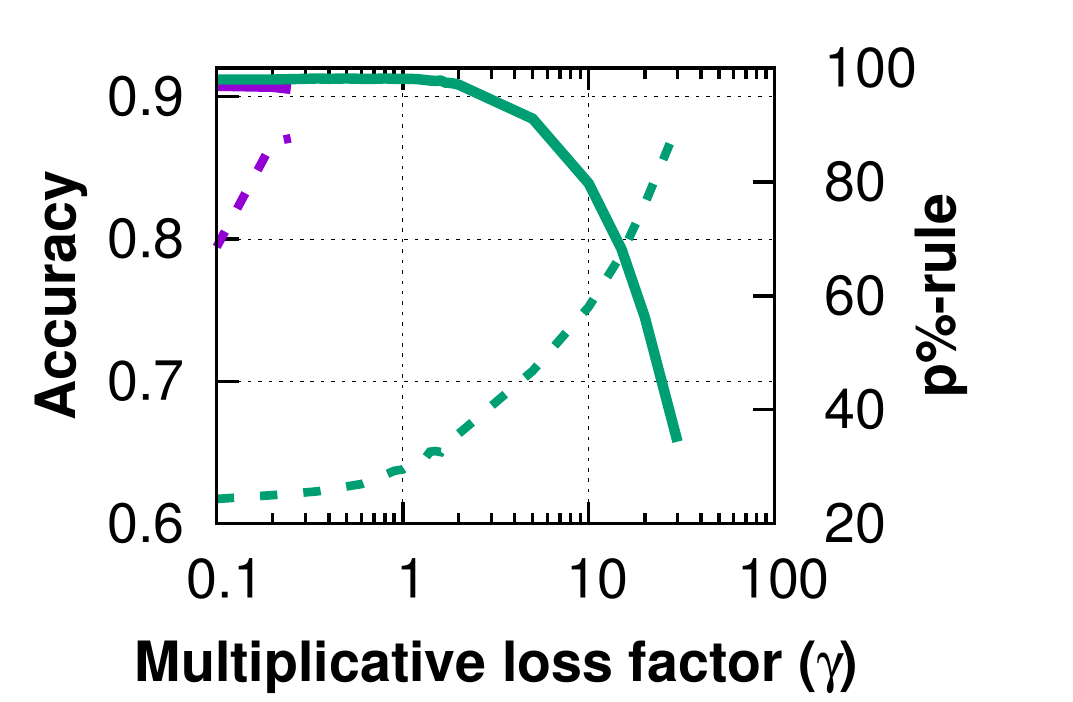} \\ \\ [-3mm]
	\small{Adult} & \small{Bank}
	\end{tabular}
	}
	\subfloat[\% in +ve class]{
	\label{frac_pos_second}
	\begin{tabular}{cc} \\
	\multicolumn{2}{c}{\includegraphics[width=.035\textwidth, angle=-90, trim = 0cm 0.5cm 0.2cm 0.5cm]{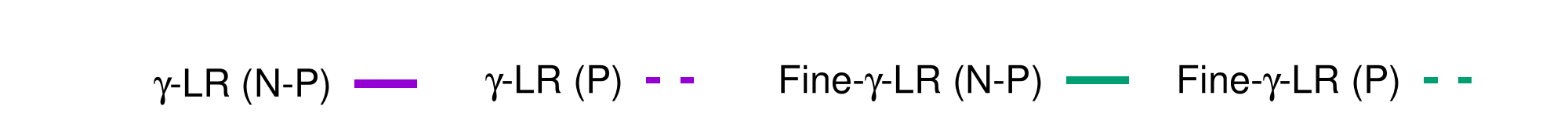}} \\ [-3mm]
	\includegraphics[width=.13\textwidth, angle=-90, trim = 0.2cm 0.5cm 0.2cm 0.5cm]{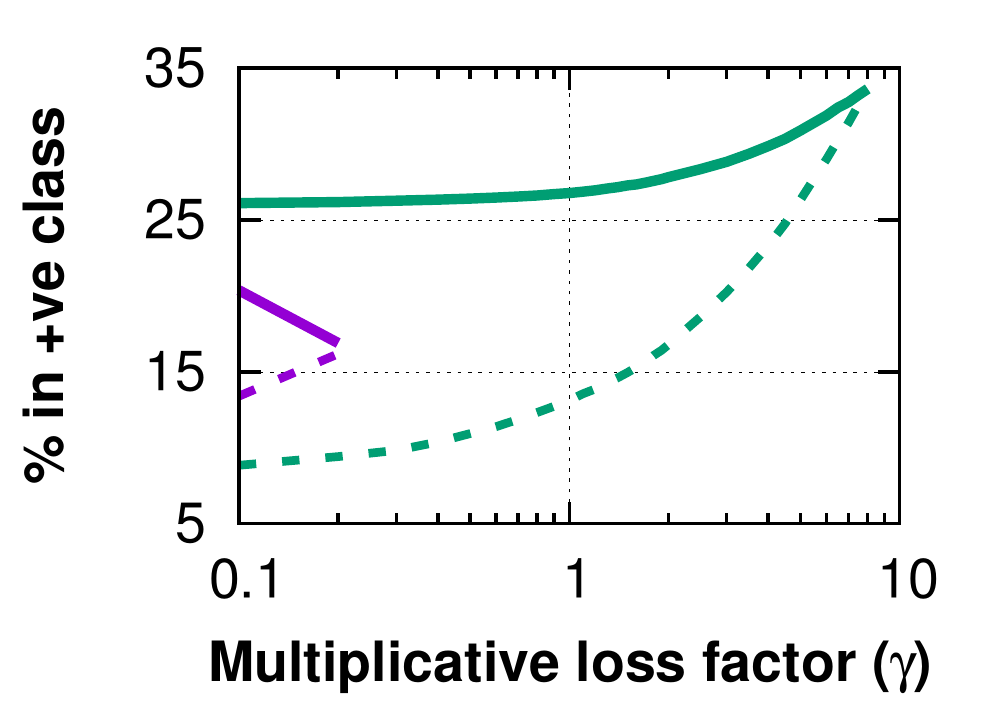} &
	\includegraphics[width=.13\textwidth, angle=-90, trim = 0.2cm 0.5cm 0.2cm 0.5cm]{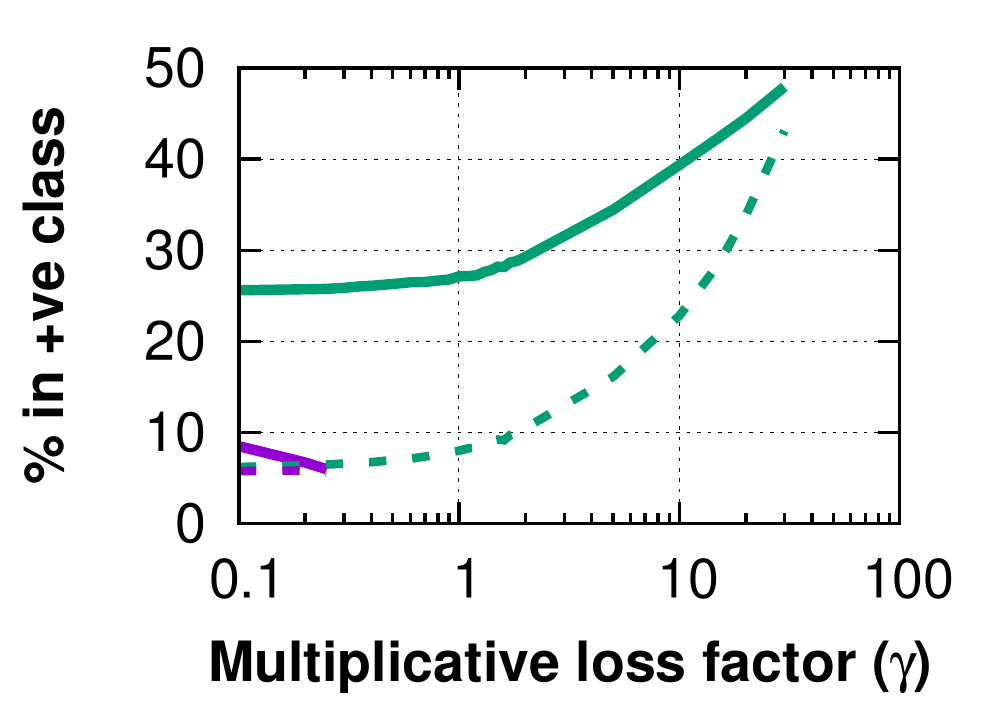} \\ \\ [-3mm]
	\small{Adult} & \small{Bank}
	\end{tabular}
	}
	\vspace{-3mm}
	\caption{[Maximizing fairness under accuracy constraints] Panels in (a) show accuracy (solid) and $p$\%-rule value (dashed) against $\gamma$.
	Panels in (b) show the percentage of protected (P, dashed) and non-protected (N-P, solid) users in the positive class against $\gamma$.} \label{fig:second-formulation}
	\vspace{-4mm}
\end{figure*}

\xhdr{Maximizing accuracy under fairness constraints}
First, we experiment with a \textbf{single binary} sensitive attribute, gender and age, for respectively, the Adult and Bank data.
For each dataset, we train several logistic regression and SVM classifiers (denoted by `C-LR' and `C-SVM', respectively), each subject to fairness constraints with different values of covariance threshold
$\mathbf{c}$ (Section~\ref{constraints}), and
then empirically investigate the trade-off between accuracy and fairness.
Fig.~\ref{fig:pareto} shows the (empirical) decision boundary covariance against the relative loss incurred by the classifier.
The `relative loss' is normalized between the loss incurred by an unconstrained classifier and by the classifier with a covariance threshold of 0.
 Here, each pair of (covariance, loss)
values is guaranteed to be Pareto optimal, since our problem formulation is convex.
Additionally, Fig.~\ref{fig:p-rule_vs_cross-cov} investigates the correspondence between decision boundary covariance and $p$\%-rule computed on the training set, showing that, as
desired: i) the lower the covariance, the higher the $p$\%-rule the classifiers satisfy and (ii) a 100\%-rule maps to zero covariance.

Then, we compare our approach to a well-known competing method from each of the two categories discussed in Section \ref{sec:intro}: preferential sampling approach~\citep{kamiran_sampling}, applied to logistic regression
(`PS-LR') and SVM (`PS-SVM'), as an example of data pre-processing, and the regularized logistic regression (`R-LR')~\citep{kamishima_regularizer}, as an example of modifying a classifier to limit unfairness.
Fig.~\ref{fig:accuracy-all} summarizes the results: the top panel shows average accuracy and the bottom panel the percentage of protected (dashed lines) and non-protected (solid lines) users in positive
class against the average $p$\%-rule, as computed on test sets.
We observe that: i) the performance of our classifiers (C-LR, C-SVM) and regularized logistic regression (R-LR) is comparable, ours are slightly better for Adult data (left column) while slightly worse for Bank data
(right column). However, R-LR uses sensitive attribute values from the test set to make predictions, failing the disparate treatment test and potentially allowing for reverse discrimination~\citep{ricci_scotus}; ii)
the preferential sampling presents the worst performance and always achieves $p$\%-rules under 80\%; and,
(iii) in the Adult data, all classifiers move non-protected users (males) to the negative class and protected users (females) to the positive class to achieve fairness, in contrast, in the Bank data, they only move
non-protected (young and old) users originally labeled as positive to the negative class since it provides a smaller accuracy loss.
However, the latter can be problematic: from a business perspective, a bank may be interested in finding potential subscribers rather than losing existing customers. This last observation motivates our
second formulation (Section~\ref{classification_accuracy_bound}), which we experiment with later in this section.

Finally, we experiment with \textbf{non-binary} (race) and \textbf{several} (gender and race) sensitive attributes  in  Adult dataset. We do not compare with competing
methods since they cannot handle non-binary or several sensitive attributes. Fig.~\ref{non_bin_several_first} summarizes the results by showing the accuracy and the percentage
of subjects sharing each sensitive attribute value classified as positive against a multiplicative covariance factor $a \in [0,1]$ such that $\mathbf{c} = a \mathbf{c}^{*}$, where $\mathbf{c}^{*}$
is the unconstrained classifier covariance\footnote{For several sensitive features, we compute the initial covariance $\mathbf{c}^{*}_{k}$ for each of the sensitive feature $k$, and then compute the covariance threshold separately for each sensitive feature as $a\mathbf{c}^{*}_{k}$.}  (note that  $p$\%-rule is only defined for a binary sensitive feature).
As expected, as the value of $\mathbf{c}$ decreases, the percentage of subjects in the positive class from sensitive attribute value groups become nearly equal
\footnote{The scarce representation of the race value `Other' (only 0.8\% of the data) hinders an accurate estimation of the decision boundary covariance and, as a result, the classifier does not reach perfect fairness with respect to this sensitive attribute value.}
while the loss in accuracy is modest.

\xhdr{Maximizing fairness under accuracy constraints}
Next, we demonstrate that our second formulation (Section~\ref{classification_accuracy_bound}) can maximize fairness while precisely controlling  loss in accuracy.
To this end,
we first train several logistic regression classifiers (denoted by `$\gamma$-LR'), which minimize the decision boundary covariance subject to accuracy constraints over the entire dataset by solving
problem~\eqref{eq:ml-no-discrimination-accuracy-bound} with increasing values of $\gamma$.
Then, we train  logistic regression classifiers (denoted by `Fine-$\gamma$-LR') that minimize the decision boundary covariance subject to \textit{fine-grained} accuracy constraints by solving  problem~\eqref{eq:ml-no-discrimination-accuracy-set}.
Here, we prevent the non-protected users that were classified as positive by the unconstrained logistic regression classifier from being classified as negative by constraining that their distance from decision boundary stays positive while learning the fair boundary.
 We then
increase $\gamma_i = \gamma$ for the remaining users.
In both cases, we increased the value of $\gamma$ until we reach a $100$\%-rule during training.
Fig.~\ref{fig:second-formulation} summarizes the results for both datasets, by showing\- (a) the average accuracy (solid curves) and $p$\%-rule (dashed curves) against $\gamma$, and (b) the percentage of non-protected (N-P, solid curves) and protected (P, dashed curves) users in the positive class against $\gamma$.
We observe that, as we increase $\gamma$, the classifiers that constrain the overall training loss ($\gamma$-LR) remove non-protected users from the positive class and add protected users to the positive class,
in contrast, the classifiers that prevent the non-protected users that were classified as positive in the unconstrained classifier from being classified as negative (Fine-$\gamma$-LR) add both protected and non-protected users to the positive class.
As a consequence, the latter achieves lower accuracy for the same $p$\%-rule.

\vspace{-4mm}
\section {DISCUSSION \& FUTURE WORK}\label{conclusions}
\vspace{-5mm}

In this paper, we introduced a novel measure of decision boundary fairness, which enables us to ensure fairness with respect to one or more sensitive attributes,  in terms of both disparate treatment
and disparate impact.
We leverage this measure to derive two complementary formulations: one that maximizes accuracy subject to fairness constraints, and helps ensure compliance with a non-discrimination policy
or law (\eg, a given $p\%$-rule); and another one that maximizes fairness subject to accuracy constraints, and ensures fulfilling certain business needs (\eg, disparate impact's business necessity clause).

Our framework opens many avenues for future work. For example, one could include fairness constraints in other supervised  (\eg, regression, recommendation)
as well as unsupervised (\eg, set selection, ranking) learning tasks.
Further, while we note that a decreasing covariance threshold corresponds to an increasing (more fair) $p$\%-rule, the relation between the two is only empirically observed. A precise mapping between covariance and $p$\%-rule is quite challenging to derive analytically since it depends on the specific classifier and the dataset being used. Such a theoretical analysis would be an interesting future direction.
Finally, in this paper we consider disparate impact as our notion of fairness with the assumption that the historical training data may contain  biases against certain group(s). Since the actual proportions in positive class from different groups (\eg, males, females) in the \text{ground-truth} dataset are not available (we only have access to a biased dataset), ensuring equal proportions from each group in the positive class (removing disparate impact) serves as an attractive notion of fairness. However, in cases where the historical ground truth decisions are available, disparate impact
can be  explained by the means of the ground truth, and alternative notions of fairness, \eg, disparate mistreatment~\citep{zafar_dmt}, might be more suitable.

\bibliographystyle{abbrvnat}

\begin{thebibliography}{24}
\providecommand{\natexlab}[1]{#1}
\providecommand{\url}[1]{\texttt{#1}}
\expandafter\ifx\csname urlstyle\endcsname\relax
  \providecommand{\doi}[1]{doi: #1}\else
  \providecommand{\doi}{doi: \begingroup \urlstyle{rm}\Url}\fi

\bibitem[{Adult data}(1996)]{adult_dataset}
{Adult data}.
\newblock \url{http://tinyurl.com/UCI-Adult}, 1996.

\bibitem[{Bank data}(2014)]{bank_dataset}
{Bank data}.
\newblock \url{http://tinyurl.com/UCI-Bank}, 2014.

\bibitem[Barocas and Selbst(2016)]{barocas_2016}
S.~Barocas and A.~D. Selbst.
\newblock {Big Data's Disparate Impact}.
\newblock \emph{{California Law Review}}, 2016.

\bibitem[Bhandari(2016)]{bhandari_aclu}
E.~Bhandari.
\newblock {Big Data Can Be Used To Violate Civil Rights Laws, and the FTC
  Agrees}.
\newblock
  \url{https://www.aclu.org/blog/free-future/big-data-can-be-used-violate-civil-rights-laws-and-ftc-agrees},
  2016.

\bibitem[Biddle(2005)]{2005adverse}
D.~Biddle.
\newblock \emph{Adverse Impact and Test Validation: A Practitioner's Guide to
  Valid and Defensible Employment Testing}.
\newblock Gower, 2005.

\bibitem[Calders and Verwer(2010)]{cadlers_naivebayes}
T.~Calders and S.~Verwer.
\newblock {Three Naive Bayes Approaches for Discrimination-Free
  Classification}.
\newblock \emph{{Data Mining and Knowledge Discovery}}, 2010.

\bibitem[{Civil Rights Act}(1964)]{civil_rights_act}
{Civil Rights Act}.
\newblock Civil Rights Act of 1964, Title VII, Equal Employment Opportunities,
  1964.

\bibitem[Dauphin et~al.(2014)Dauphin, Pascanu, Gulcehre, Cho, Ganguli, and
  Bengio]{Dauphin_2014}
Y.~N. Dauphin, R.~Pascanu, C.~Gulcehre, K.~Cho, S.~Ganguli, and Y.~Bengio.
\newblock Identifying and {A}ttacking the {S}addle {P}oint {P}roblem in
  {H}igh-dimensional {N}on-convex {O}ptimization.
\newblock In \emph{NIPS}, 2014.

\bibitem[Dwork et~al.(2012)Dwork, Hardt, Pitassi, and Reingold]{Dwork2012}
C.~Dwork, M.~Hardt, T.~Pitassi, and O.~Reingold.
\newblock Fairness {T}hrough {A}wareness.
\newblock In \emph{ITCSC}, 2012.

\bibitem[Feldman et~al.(2015)Feldman, Friedler, Moeller, Scheidegger, and
  Venkatasubramanian]{feldman_kdd15}
M.~Feldman, S.~A. Friedler, J.~Moeller, C.~Scheidegger, and
  S.~Venkatasubramanian.
\newblock Certifying and removing disparate impact.
\newblock In \emph{KDD}, 2015.

\bibitem[Goh et~al.(2016)Goh, Cotter, Gupta, and Friedlander]{goh_nips2016}
G.~Goh, A.~Cotter, M.~Gupta, and M.~Friedlander.
\newblock {Satisfying Real-world Goals with Dataset Constraints}.
\newblock In \emph{{NIPS}}, 2016.

\bibitem[Kamiran and Calders(2009)]{kamiran_classifying}
F.~Kamiran and T.~Calders.
\newblock {Classifying without Discriminating}.
\newblock In \emph{{IC4}}, 2009.

\bibitem[Kamiran and Calders(2010)]{kamiran_sampling}
F.~Kamiran and T.~Calders.
\newblock {Classification with No Discrimination by Preferential Sampling}.
\newblock In \emph{{BENELEARN}}, 2010.

\bibitem[Kamishima et~al.(2011)Kamishima, Akaho, Asoh, and
  Sakuma]{kamishima_regularizer}
T.~Kamishima, S.~Akaho, H.~Asoh, and J.~Sakuma.
\newblock {Fairness-aware Classifier with Prejudice Remover Regularizer}.
\newblock In \emph{{PADM}}, 2011.

\bibitem[Luong et~al.(2011)Luong, Ruggieri, and Turini]{salvatore_knn}
B.~T. Luong, S.~Ruggieri, and F.~Turini.
\newblock {kNN as an Implementation of Situation Testing for Discrimination
  Discovery and Prevention}.
\newblock In \emph{{KDD}}, 2011.

\bibitem[Mu{\~n}oz et~al.(2016)Mu{\~n}oz, Smith, and
  Patil]{bigdatawhitehouse2016}
C.~Mu{\~n}oz, M.~Smith, and D.~Patil.
\newblock {Big Data: A Report on Algorithmic Systems, Opportunity, and Civil
  Rights}.
\newblock \emph{Executive Office of the President. The White House.}, 2016.

\bibitem[Pedreschi et~al.(2008)Pedreschi, Ruggieri, and
  Turini]{pedreschi_discrimination}
D.~Pedreschi, S.~Ruggieri, and F.~Turini.
\newblock {Discrimination-aware Data Mining}.
\newblock In \emph{{KDD}}, 2008.

\bibitem[Podesta et~al.(2014)Podesta, Pritzker, Moniz, Holdren, and
  Zients]{bigdatawhitehouse2014}
J.~Podesta, P.~Pritzker, E.~Moniz, J.~Holdren, and J.~Zients.
\newblock Big {D}ata: {S}eizing {O}pportunities, {P}reserving {V}alues.
\newblock \emph{Executive Office of the President. The White House.}, 2014.

\bibitem[{Ricci vs. DeStefano}(2009)]{ricci_scotus}
{Ricci vs. DeStefano}.
\newblock {U.S. Supreme Court}, 2009.

\bibitem[Romei and Ruggieri(2014)]{salvatore_survey}
A.~Romei and S.~Ruggieri.
\newblock {A Multidisciplinary Survey on Discrimination Analysis}.
\newblock \emph{{KER}}, 2014.

\bibitem[Sch{\"o}lkopf and Smola(2002)]{scholkopf2002learning}
B.~Sch{\"o}lkopf and A.~J. Smola.
\newblock \emph{Learning with Kernels: Support Vector Machines, Regularization,
  Optimization, and Beyond}.
\newblock MIT press, 2002.

\bibitem[Sweeney(2013)]{sweeney_queue}
L.~Sweeney.
\newblock {Discrimination in Online Ad Delivery}.
\newblock \emph{{ACM Queue}}, 2013.

\bibitem[Zafar et~al.(2017)Zafar, Valera, Rodriguez, and Gummadi]{zafar_dmt}
M.~B. Zafar, I.~Valera, M.~G. Rodriguez, and K.~P. Gummadi.
\newblock {Fairness Beyond Disparate Treatment \& Disparate Impact: Learning
  Classification without Disparate Mistreatment}.
\newblock In \emph{{WWW}}, 2017.

\bibitem[Zemel et~al.(2013)Zemel, Wu, Swersky, Pitassi, and
  Dwork]{icml2013_zemel13}
R.~Zemel, Y.~Wu, K.~Swersky, T.~Pitassi, and C.~Dwork.
\newblock Learning {F}air {R}epresentations.
\newblock In \emph{ICML}, 2013.

\end{thebibliography}

\newpage
\begin{appendix}

\onecolumn
\clearpage
\normalsize

\section{Particularizing the Fairness Constraints for SVM} \label{sec:svm_formulation}

One can specialize our fair classifier formulation proposed in~\eqref{eq:general-classifier-no-discrimination} as:

\xhdr{Linear SVM}
A linear SVM distinguishes among classes using  a linear hyperplane $\thetab^{T} \mathbf{x} = 0$. In this case, the parameter vector $\thetab$ of the \emph{fair} linear SVM can be found by solving the following quadratic program:
\begin{align}
       \begin{array}{ll} \nonumber
        \mbox{minimize} & \| \thetab \|^2 + C \sum_{i=1}^n  \xi_i  \\
    \mbox{subject to} &  y_i\thetab^{T}  \mathbf{x}_i \geq 1- \xi_i ,   \forall i \in \{1, \ldots, n\}\\
    & \xi_i \geq  0, \forall i \in \{1, \ldots, n\}, \\
    \end{array} &
     \\
     \begin{array}{ll}
       & \frac{1}{N} \sum_{i=1}^{N} \left(\mathbf{z}_i - \bar{\mathbf{z}}\right)\thetab^{T}  \mathbf{x}_i \leq \mathbf{c},   \\
       &  \frac{1}{N} \sum_{i=1}^{N} \left(\mathbf{z}_i - \bar{\mathbf{z}}\right) \thetab^{T}  \mathbf{x}_i \geq -\mathbf{c}, \hspace{2mm} \quad   \quad
    \end{array} &   \label{eq:SVM-no-discrimination}
\end{align}
where $\thetab$ and $\xi$ are the variables, $\| \thetab \|^2$ corresponds to the margin between the \emph{support vectors} assigned to different classes, and $C \sum_{i=1}^n  \xi_i $ penalizes the number of data points falling inside the margin.

\xhdr{Nonlinear SVM}
In a nonlinear SVM, the decision boundary takes the form $\thetab^{T} \Phi(\mathbf{x}) = 0$, where $\Phi(\cdot)$ is a nonlinear transformation that maps every feature vector $\mathbf{x}$ into a higher dimensional transformed feature space. Similarly
as in the case of a linear SVM, one may think of finding the parameter vector $\thetab$ by solving a constrained quadratic program similar to the one defined by
Eq.~\eqref{eq:SVM-no-discrimination}.
However, the dimensionality of the transformed feature space can be large, or even infinite, making the corresponding optimization problem difficult to solve.
Fortunately, we can leverage the \emph{kernel trick}~\citep{scholkopf2002learning} both in the original optimization problem and the fairness inequalities, and resort instead to the dual form of the problem, which can be
solved efficiently. In particular, the dual form is given by:
\begin{align}
       \begin{array}{ll} \nonumber
                \mbox{minimize} &  \sum_{i=1}^N   \alpha_i + \sum_{i=1}^N \alpha_i  y_i (g_{\alphab}(\mathbf{x}_i) + h_{\alphab}(\mathbf{x}_i)) \\
        \mbox{subject to} & \alpha_i \geq 0 ,   \forall i \in \{1, \ldots, N\},\\
        & \sum_{i=1}^N  \alpha_i  y_i= 0, \\
    \end{array} &  \\
     \begin{array}{ll}
       &  \frac{1}{N} \sum_{i=1}^{N}  \left(\mathbf{z}_i - \bar{\mathbf{z}}\right) g_{\alphab}(\mathbf{x}_i) \leq \mathbf{c},   \\
       &  \frac{1}{N} \sum_{i=1}^{N} \left(\mathbf{z}_i - \bar{\mathbf{z}}\right) g_{\alphab}(\mathbf{x}_i) \geq -\mathbf{c}, \hspace{11mm} \quad   \quad
    \end{array} &   \label{eq:SVM-nonlinear-dual}
\end{align}
where $\alphab$ are the dual variables, $g_{\alphab}(\mathbf{x}_i) = \sum_{j=1}^N  \alpha_j y_j k(  \mathbf{x}_i , \mathbf{x}_j )$ can still be interpreted as a signed distance to the decision boundary in
the transformed feature space, and $h_{\alphab}(\mathbf{x}_i) = \sum_{j=1}^N  \alpha_j y_j \frac{1}{C} \delta_{ij}$, where $\delta_{ij} = 1$ if $i=j$ and $\delta_{ij} = 0$, otherwise.  Here, $k(\mathbf{x}_i, \mathbf{x}_j) = \langle \phi(\mathbf{x}_i), \phi(\mathbf{x}_j) \rangle$ denotes the inner product between a pair of transformed feature vectors and is often called the kernel function.

\section{Additional Experiments} \label{app:results}

\subsection{Experiments on Non-linear Synthetic Data}\label{app:syn_exp}

Here, we illustrate how the decision boundary of an non-linear classifier, a SVM with radial basis function (RBF) kernel, changes under fairness constraints.
To this end, we generate $4$,$000$ user binary class labels uniformly at random and assign a 2-dimensional user feature vector per label by drawing samples
from
$p(x| y=1, \beta) = \beta N([2; 2], [5\,\, 1; 1\,\, 5]) +  (1-\beta) N([-2; -2], [10\,\, 1; 1\,\, 3]) $ if $y = 1$,
and $p(x| y=-1, \beta ) = \beta N([4; -4], [4\,\, 4; 2\,\, 5]) +  (1-\beta) N([-4; 6], [6\,\, 2; 2\,\, 3])$ otherwise, where $\beta \in\{0,1\}$ is sampled from $Bernoulli (0.5)$.
Then, we generate each user'{}s sensitive attri\-bute $z$ by applying the same rotation as detailed in Section~\ref{sec:synthetic}.

\begin{figure*}[th]
	\centering

	\subfloat[Unconstrained]{\includegraphics[width=.3\textwidth, trim = 2cm 1.5cm 0.0cm 0cm]{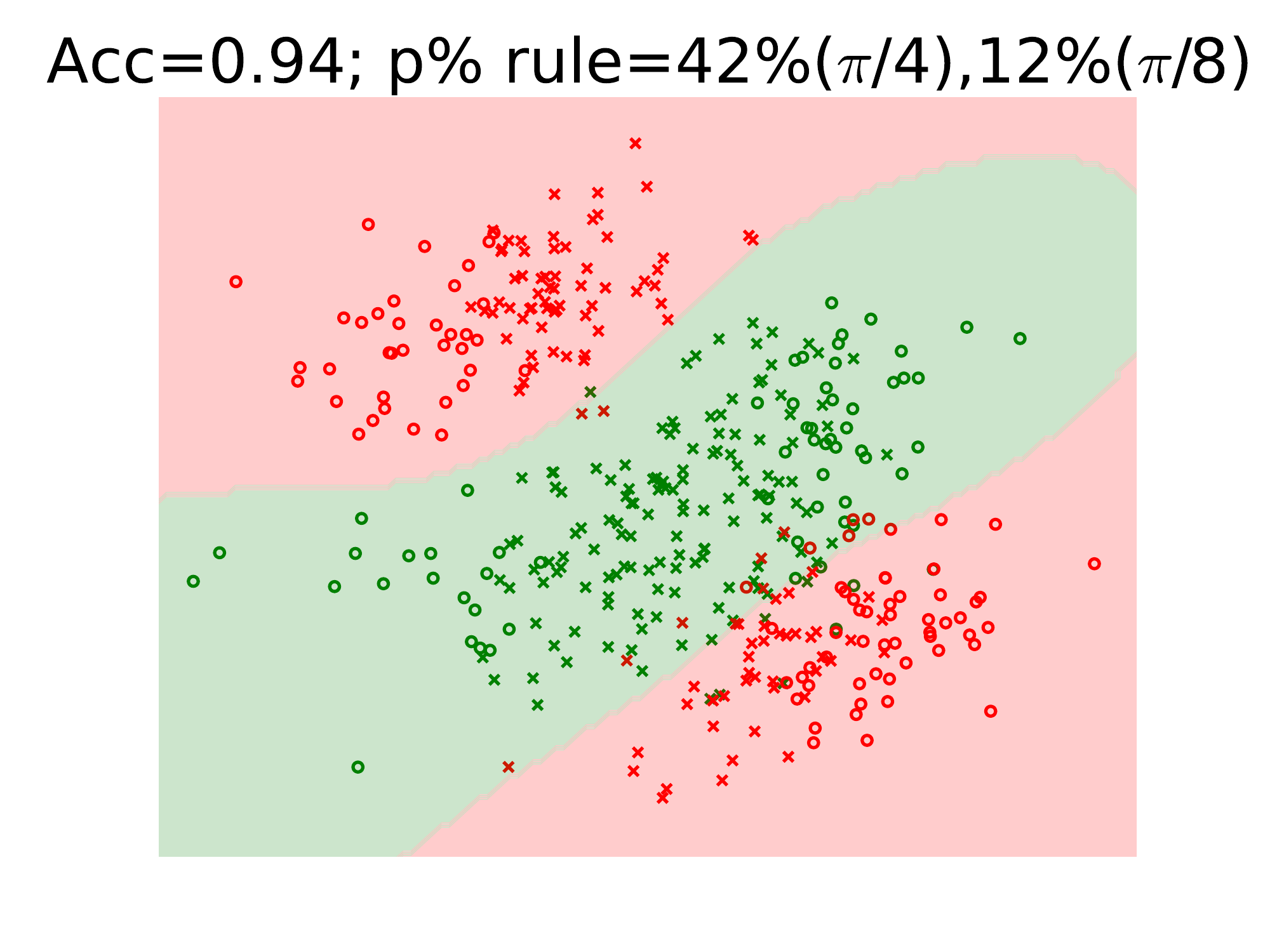}} \hspace{5mm}
	\subfloat[$\phi = \pi/4$]{\includegraphics[width=.3\textwidth, trim = 2cm 1.5cm 0.0cm 0cm]{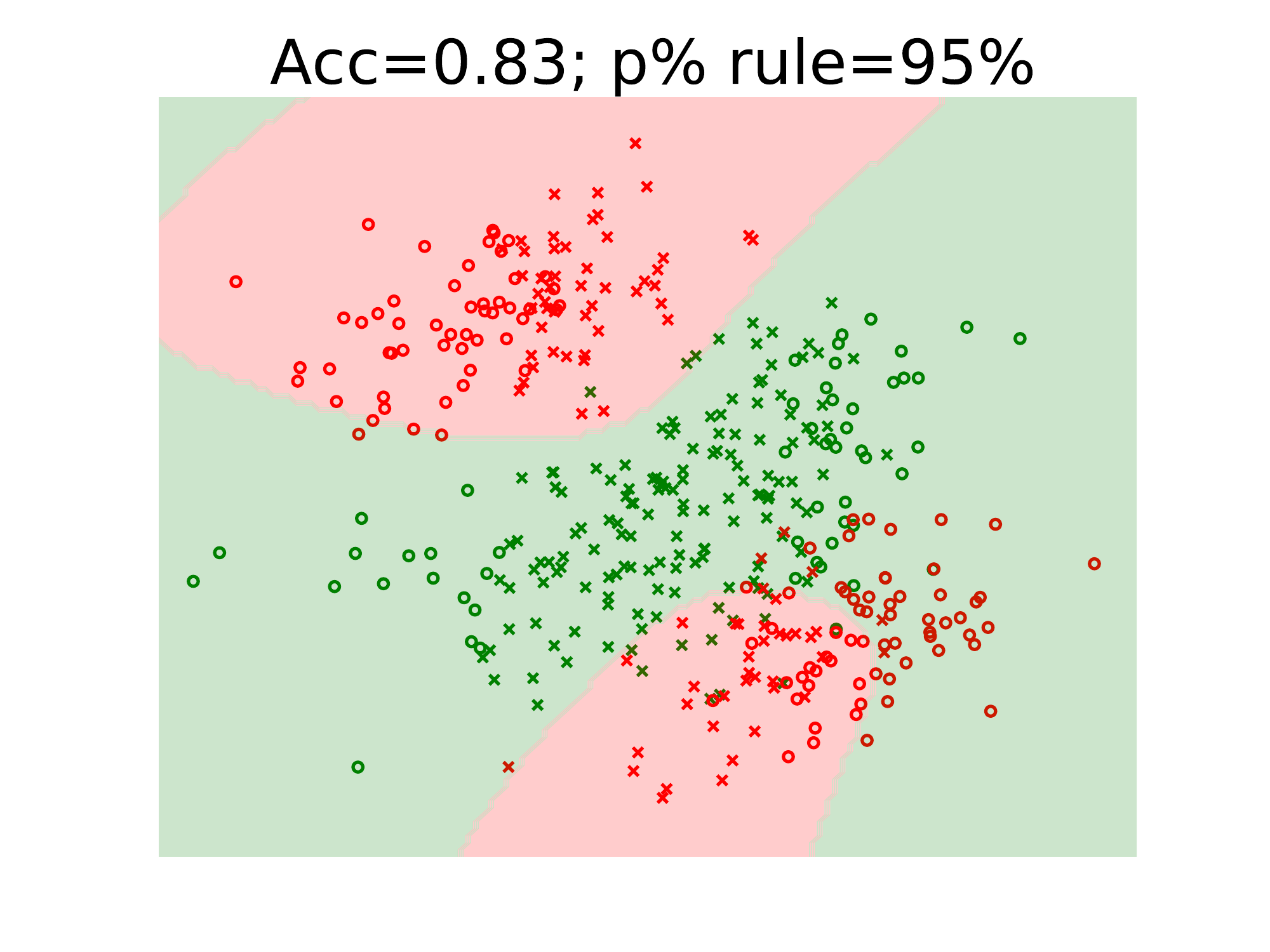}} \hspace{5mm}
	\subfloat[$\phi = \pi/8$]{\includegraphics[width=.3\textwidth, trim = 2cm 1.5cm 0.0cm 0cm]{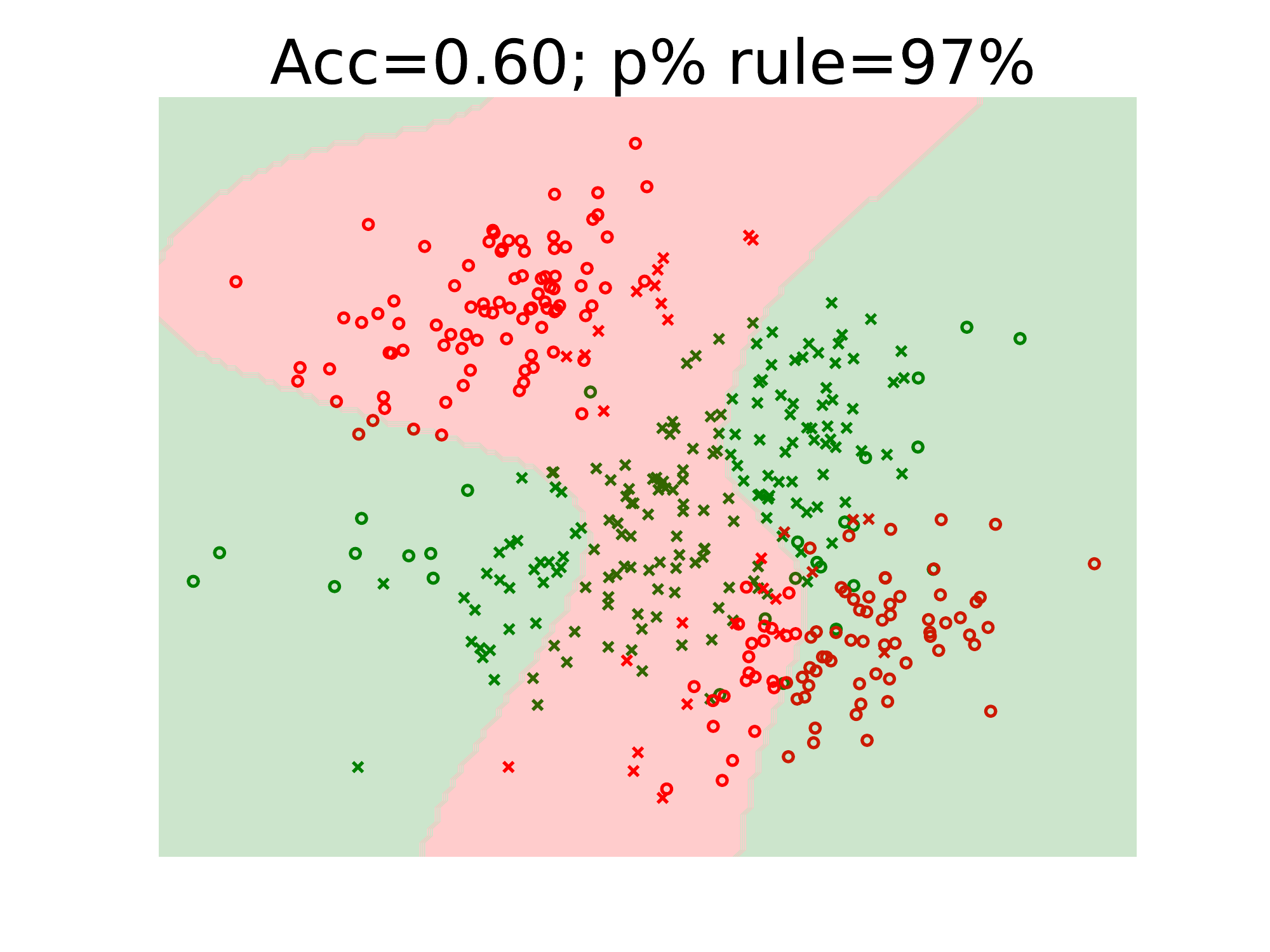}}
	\caption{Decision boundaries for SVM classifier with RBF Kernel trained without fairness constraints (left) and with fairness constraints ($\mathbf{c}=0$) on two synthetic datasets with different correlation value between sensitive attribute values (crosses vs circles) and class labels (red vs green).
	} \label{fig:decision-boundary-synthetic-non-linear}
\end{figure*}

Figure~\ref{fig:decision-boundary-synthetic-non-linear} shows the decision boundaries provided by the SVM that maximizes accuracy under fairness constraints with $\mathbf{c} = 0$ for two different
correlation values, set by $\phi = \pi/4$ and $\phi=\pi/8$, in comparison with the unconstrained SVM.
We observe that, in this case, the decision boundaries provided by the constrained SVMs are very different to the decision boundary provided by the unconstrained SVM, not simple shifts
or rotations of the latter, and successfully reverse engineer the mechanism we used to generate the class labels and sensitive attributes.

\subsection{Experiments on Real Data}\label{app:real_exp}
\normalsize
\xhdr{Additional Data Statistics}
In this section, we show the distribution of sensitive features and class labels in our real-world datasets.
\begin{table}[h]
\caption{Datasets details (binary sensitive attributes: gender and age).}
\subfloat[Adult dataset]{
\small
\begin{tabular}{|c|c|c|c|}
\hline
Sensitive Attribute  & $y \leq 50K$ & $>50K$ & Total \\ \hline
Males &  $ 20{,}988$ &  $ 9{,}539$  &  $ 30{,}527$\\
Females &  $ 13{,}026$ & $1{,}669 $  &  $14,695 $\\ \hline
Total & $34{,}014$ &  $11{,}208$
&  $ 45{,}222$ \\\hline
\end{tabular}
\label{tab:adult}}
\subfloat[Bank dataset]{
\small
\begin{tabular}{|c|c|c|c|}
\hline
Sensitive Attribute  & No & Yes & Total \\ \hline
25 $\leq$ age $\leq$ 60 &  $ 35{,}240$ &  $3{,}970 $  &  $39{,}210 $\\
age $<$ 25 or age $>$ 60 &  $1{,}308 $ & $ 670$  &  $ 1{,}978$\\ \hline
Total & $36{,}548 $ & $ 4{,}640$  &  $ 41{,}188$ \\ \hline\end{tabular}
\label{tab:bank}}
\end{table}
\begin{table}[h]
\caption{Adult dataset (Non-binary sensitive attribute: race)}
\centering
\small
\begin{tabular}{|c|c|c|c|}
\hline
Sensitive Attribute  & $y \leq 50K$ & $>50K$ & Total \\ \hline
American-Indian/Eskimo &  $ 382 $ &  $ 53$  &  $ 435$\\
Asian/Pacific-Islander &  $ 934 $ & $ 369 $  &  $1{,}303 $\\
White &  $  28{,}696$ & $10{,}207 $  &  $38,903 $\\
Black &  $ 3{,}694$ & $534 $  &  $4,228 $\\
Other &  $ 308$ & $45 $  &  $ 353 $\\ \hline
Total & $34{,}014$ &  $11{,}208$  &  $ 45{,}222$ \\ \hline
\end{tabular}
\label{tab:adult_race}
\end{table}
\normalsize

\pagebreak

\begin{figure*}[t]
\vspace{-5mm}
    \centering
    \subfloat[Loss vs. Cov.]{
    \label{fig:pareto_cv}
    \begin{tabular}{c}
    \multicolumn{1}{c}{\hspace{4mm} \includegraphics[width=.038\textwidth, angle=-90]{legend_cov_vs_p_rule}} \\ [-3mm]
    \includegraphics[width=.15\textwidth, angle=-90, trim = 0.2cm 0.5cm 0.2cm 0.5cm]{pareto_loss_adult}\\ [-3mm]
    \multicolumn{1}{c}{\hspace{4mm} \includegraphics[width=.038\textwidth, angle=-90]{legend_cov_vs_p_rule}} \\[-3mm]
    \setcounter{subfigure}{1}\includegraphics[width=.15\textwidth, angle=-90, trim = 0.2cm 0.5cm 0.2cm 0.5cm]{pareto_loss_bank}  \\ \\ [-3mm]
    \small{ \hspace{2mm} Adult and Bank}
    \end{tabular}
    }
    \hspace{-6mm}
    \subfloat[Cov. vs. CV score]{
    \label{fig:p-rule_vs_cross-cov_cv}
    \begin{tabular}{c}
    \multicolumn{1}{c}{\hspace{4mm} \includegraphics[width=.035\textwidth, angle=-90, trim = 0cm 0.5cm 0.2cm 0.5cm]{legend_cov_vs_p_rule}} \\ [-3mm]
    \includegraphics[width=.15\textwidth, angle=-90, trim = 0.2cm 0.5cm 0.2cm 0.5cm]{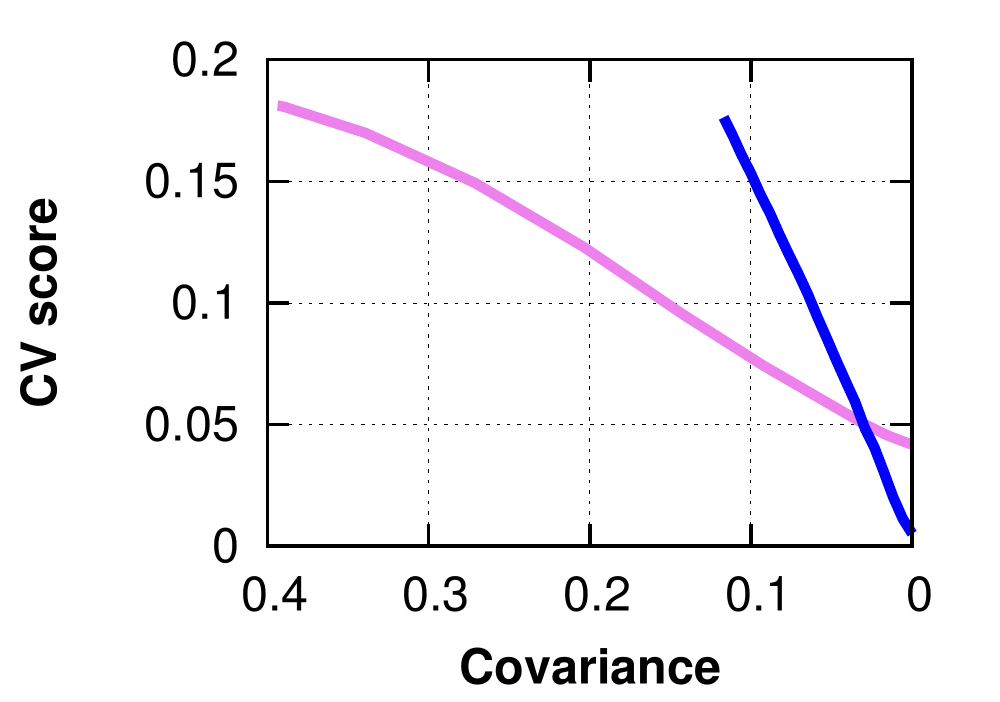}\\ [-3mm]
    \multicolumn{1}{c}{\hspace{4mm} \includegraphics[width=.035\textwidth, angle=-90, trim = 0cm 0.5cm 0.2cm 0.5cm]{legend_cov_vs_p_rule}} \\[-3mm]
    \includegraphics[width=.15\textwidth, angle=-90, trim = 0.2cm 0.5cm 0.2cm 0.5cm]{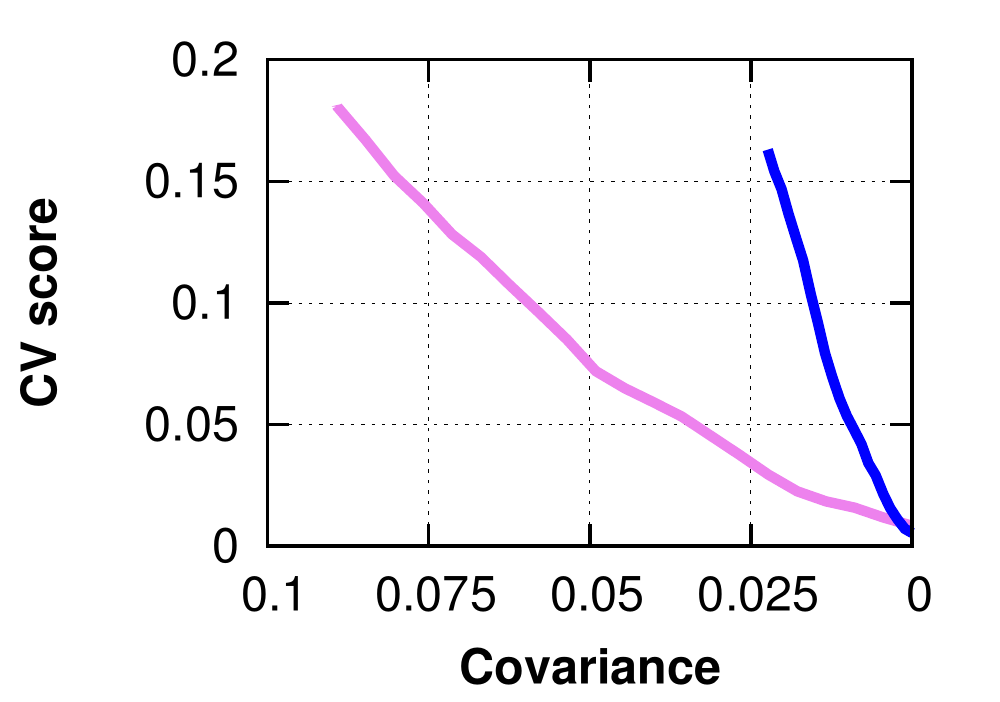}  \\ \\ [-3mm]
    \small{ \hspace{2mm} Adult and Bank}
    \end{tabular}
    }
    \subfloat[Single binary sensitive attribute]{
    \label{fig:accuracy-all-cv}
    \begin{tabular}{cc}
    \multicolumn{2}{c}{\hspace{-3mm} \includegraphics[width=.035\textwidth, angle=-90, trim = 0cm 0.5cm 0.2cm 0.5cm]{legend_p_rule}} \\ [-3mm]
    \hspace{-3mm}  \includegraphics[width=.15\textwidth, angle=-90, trim = 0.2cm 0.5cm 0.2cm 0.5cm]{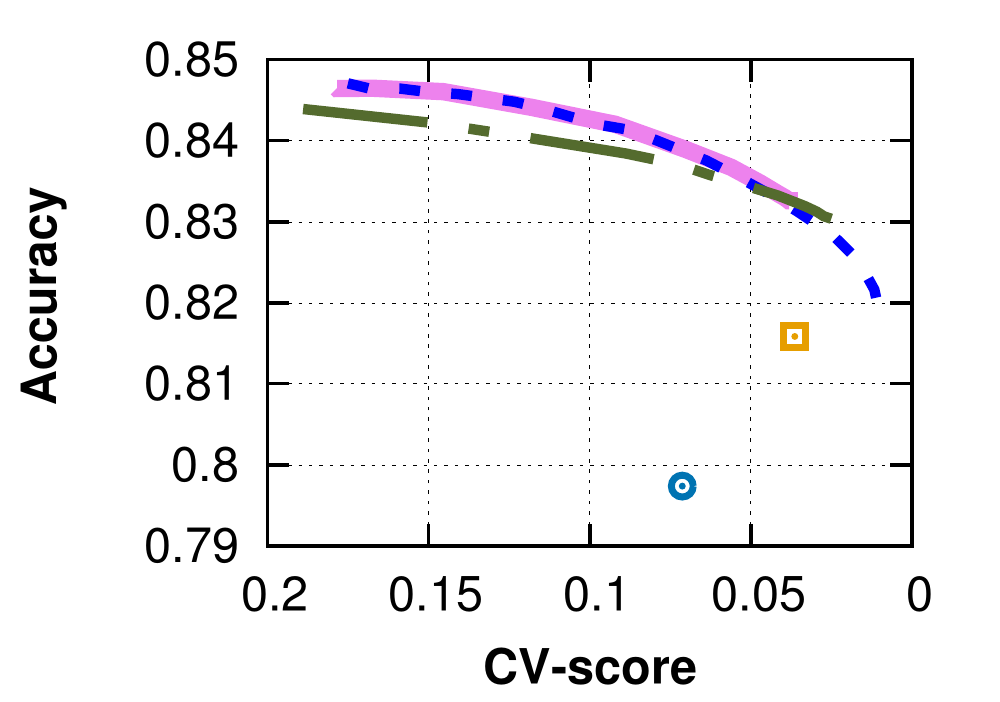}   &
     \includegraphics[width=.15\textwidth, angle=-90, trim = 0.2cm 0.5cm 0.2cm 0.5cm]{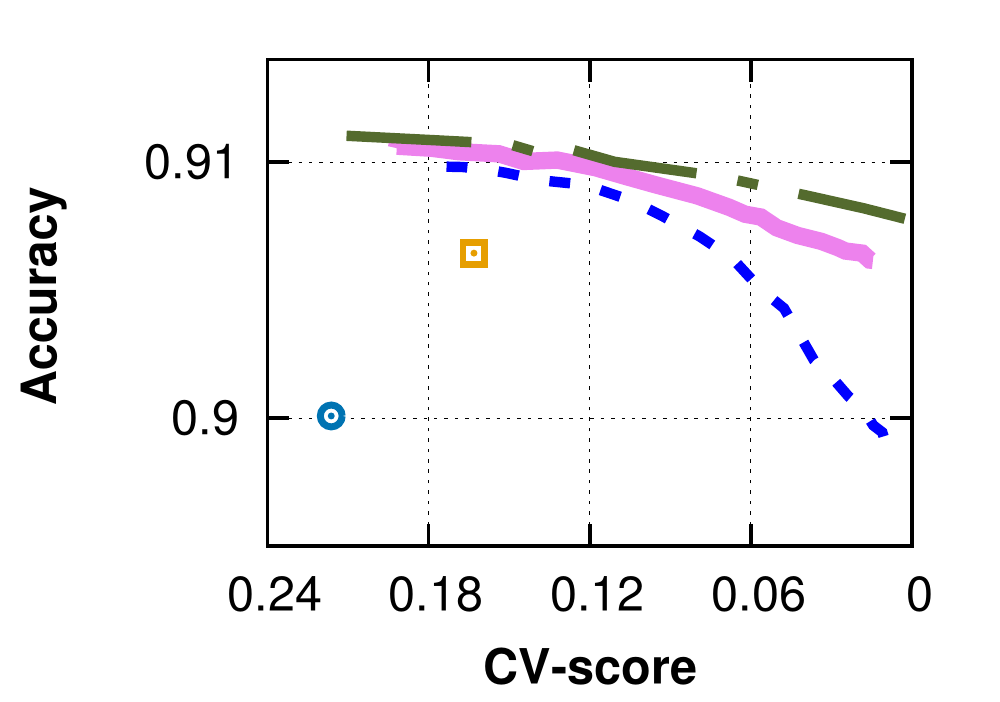} \\ [-3mm]
    \multicolumn{2}{c}{\hspace{-3mm} \includegraphics[width=.035\textwidth, angle=-90, trim = 0cm 0.5cm 0.2cm 0.5cm]{legend_frac_pos_first}}  \\  [-3mm]
    \hspace{-3mm} \includegraphics[width=.15\textwidth, angle=-90, trim = 0.2cm 0.5cm 0.2cm 0.5cm]{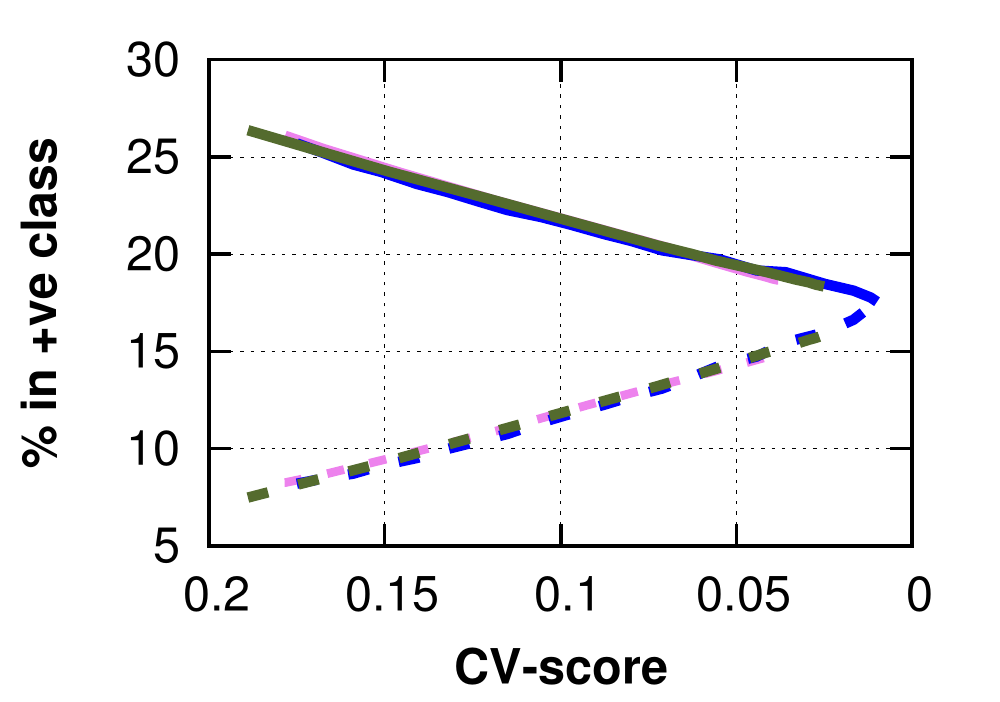}  &
    \includegraphics[width=.15\textwidth, angle=-90, trim = 0.2cm 0.5cm 0.2cm 0.5cm]{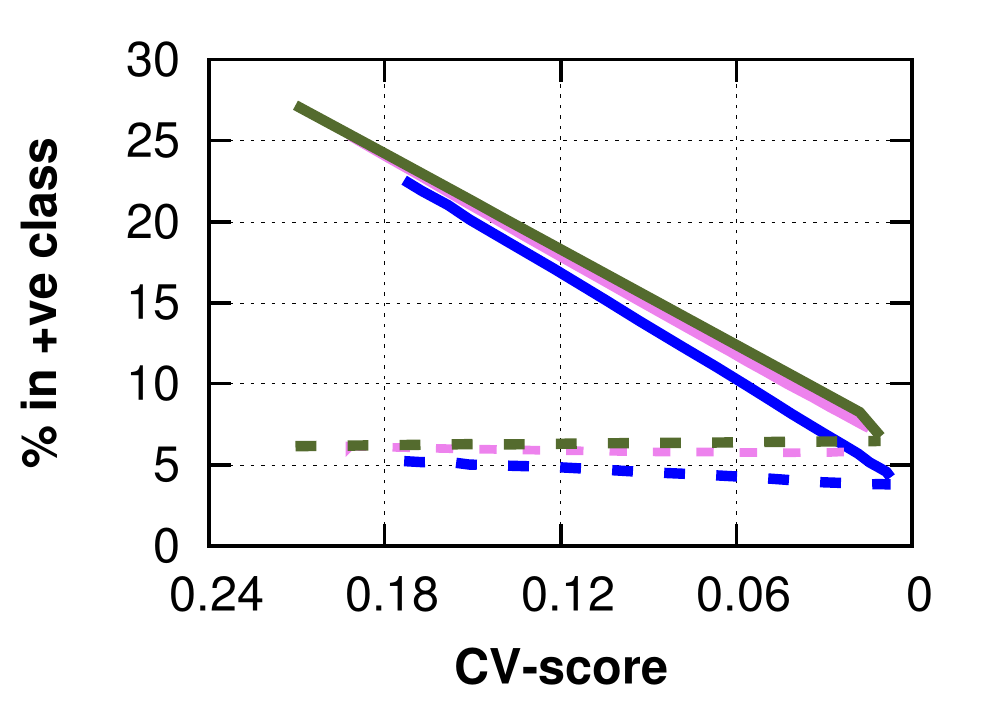} \\ \\ [-3mm]
    \small{\hspace{2mm}Adult} & \small{ \hspace{2mm}Bank}
    \end{tabular}
    }
    \vspace{-4mm}
    \caption{
    [Maximizing accuracy under fairness constraints: single, binary sensitive attribute]
        Panels in (a) show the trade-off between the empirical covariance and the relative loss in accuracy (with respect to the unconstrained classifier), where each pair of (covariance, loss)
        values is guaranteed to be Pareto optimal by construction.
        Panels in (b) show the correspondence between the empirical covariance in Eq.~\ref{eq:fairness-definition} and the CV score for classifiers trained under fairness constraints for the Adult (top) and Bank (bottom) datasets.
        Panels in (c) show the accuracy against CV score value (top) and the percentage of protected (dashed) and non-protected (solid) users in the positive class against the CV score value (bottom).
    }
    \vspace{-3mm}
\end{figure*}

\begin{figure}[H]
    \centering
    \subfloat[Acc and CV score]{
    \label{acc_and_cv_second}
    \begin{tabular}{cc} \\
    \multicolumn{2}{c}{\includegraphics[width=.04\textwidth, angle=-90, trim = 0cm 0.5cm 0.2cm 0.5cm]{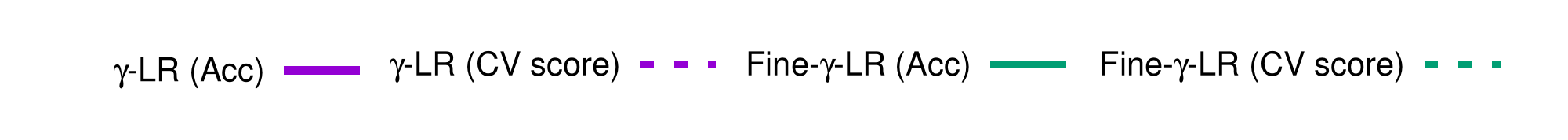}} \\ [-3mm]
    \includegraphics[width=.14\textwidth, angle=-90, trim = 0.2cm 0.5cm 0.2cm 0.5cm]{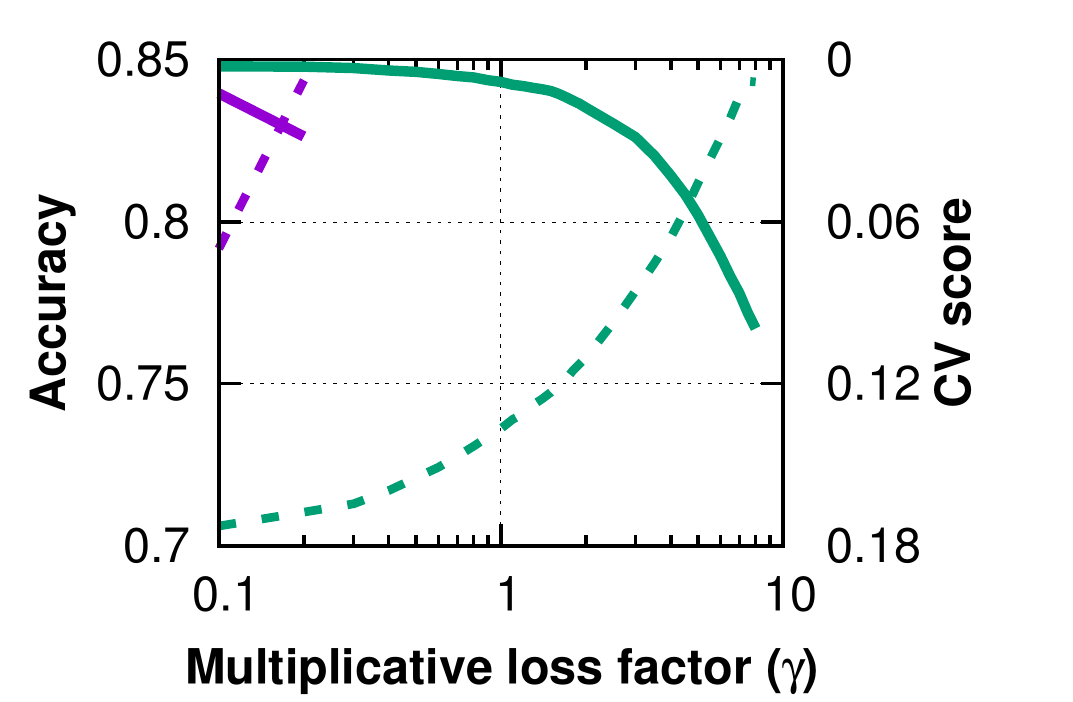} &
    \includegraphics[width=.14\textwidth, angle=-90, trim = 0.2cm 0.5cm 0.2cm 0.5cm]{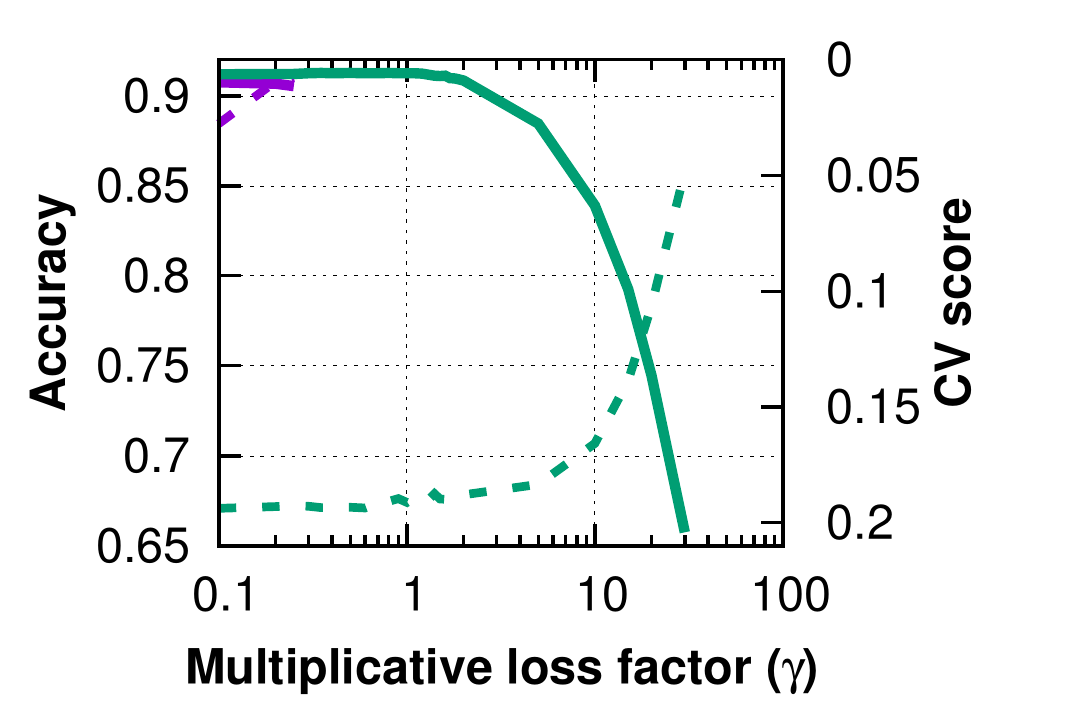} \\ \\ [-3mm]
    \small{Adult} & \small{Bank}
    \end{tabular}
    }
    \subfloat[\% in +ve class]{
    \label{frac_pos_second_cv}
    \begin{tabular}{cc} \\
    \multicolumn{2}{c}{\includegraphics[width=.04\textwidth, angle=-90, trim = 0cm 0.5cm 0.2cm 0.5cm]{legend_frac_pos_second}} \\ [-3mm]
    \includegraphics[width=.14\textwidth, angle=-90, trim = 0.2cm 0.5cm 0.2cm 0.5cm]{adult_gamma_perc_pos_plot} &
    \includegraphics[width=.14\textwidth, angle=-90, trim = 0.2cm 0.5cm 0.2cm 0.5cm]{bank_gamma_perc_pos_plot} \\ \\ [-3mm]
    \small{Adult} & \small{Bank}
    \end{tabular}
    }
    \caption{[Maximizing fairness under accuracy constraints] Panels in (a) show the accuracy (solid) and CV score value (dashed) against $\gamma$. Panels in (b) show the percentage of protected (P, dashed) and
    non-protected (N-P, solid) users in the positive class against $\gamma$.} \label{fig:second-formulation-cv}
\end{figure}

\xhdr{CV score as fairness measure}
While evaluating the performance of our method in Section~\ref{evaluation}, we used $p$\%-rule as the true measure of fairness, since it is a generalization of the $80$\%-rule advocated by US Equal Employment Opportunity Commission~\citep{2005adverse} to quantify disparate impact. We would like to remark that this measure is closely related to another measure of fairness used by some of the previous works~\citep{kamiran_classifying,kamishima_regularizer,icml2013_zemel13} in this area, referred to as Calder-Verwer (CV) score by~\citet{kamishima_regularizer}.
In particular, the CV score is defined as the (absolute value of the) difference between the percentage of users sharing a particular sensitive attribute value that lie on one side of the decision boundary and the percentage of users not sharing that value lying on the same side, \ie, $\big| P(d_{\thetab}(\mathbf{x}) \geq 0 | z=0) - P(d_{\thetab}(\mathbf{x}) \geq 0 | z=1) \big|$.

In this section, we show that using CV score (instead of $p$\%-rule) as a measure of fairness would yield similar results.

First, we show that constraining the covariance between users' sensitive attributes (Fig. \ref{fig:pareto_cv} and Figure \ref{fig:p-rule_vs_cross-cov_cv}), and the signed distance from the decision boundary, corresponds to an increasing relative loss and decreasing CV score (a more fair decision boundary).

Next, we show the performance of different methods based on the CV score (Fig, \ref{fig:accuracy-all-cv} and \ref{fig:second-formulation-cv}).
The results in Fig.\ref{fig:accuracy-all-cv} and \ref{fig:second-formulation-cv} correspond to the ones shown in Fig.  \ref{fig:accuracy-all} and \ref{fig:second-formulation}, where we took $p$\%-rule as the measure of fairness. It can be seen that both measures of fairness ($p$\%-rule and CV score) provide very similar trades-off in terms of fairness and accuracy.

Notice that according to the definitions provided in Section \ref{sec:definition}, a decreasing CV score corresponds to an increasing $p$\%-rule (and hence, a more fair decision boundary).

\makeatletter
\setlength{\@fptop}{0pt}
\makeatother

\end{appendix}

\end{document}